\pdfoutput=1
\documentclass{IEEEtran}

\makeatletter
\def\endthebibliography{%
	\def\@noitemerr{\@latex@warning{Empty `thebibliography' environment}}%
	\endlist
}
\makeatother

\usepackage{cite}
\usepackage{graphicx}
\usepackage{textcomp}
\def\BibTeX{{\rm B\kern-.05em{\sc i\kern-.025em b}\kern-.08em
    T\kern-.1667em\lower.7ex\hbox{E}\kern-.125emX}}

\usepackage[noend]{algorithmic}
\usepackage{algorithm}
\usepackage{microtype}
\usepackage{graphicx}
\usepackage{subfigure}
\usepackage{booktabs}
\usepackage{hyperref}
\usepackage{color}

\usepackage{amssymb}
\usepackage{amsmath}
\usepackage{verbatim}
\usepackage{bm}
\usepackage{amsmath}
\usepackage{amsfonts}
\usepackage{multirow}

\newtheorem{Definition}{Definition}
\newtheorem{theorem}{Theorem}

\DeclareMathOperator*{\Exp}{\mathbb{E}}

\begin{document}

\newcommand{\fracpartial}[2]{\frac{\partial #1}{\partial  #2}}
\newcommand{\norm}[1]{\left\lVert#1\right\rVert}
\newcommand{\innerproduct}[2]{\left\langle#1, #2\right\rangle}
\newcommand{\fan}[1]{\Vert #1 \Vert}
\newcommand{\qileft}{[\kern-0.15em[}
\newcommand{\qiLeft}{\left[\kern-0.4em\left[}
\newcommand{\qiright}{]\kern-0.15em]}
\newcommand{\qiRight}{\right]\kern-0.4em\right]}
\newcommand{\sign}{{\mbox{sign}}}
\newcommand{\diag}{{\mbox{diag}}}
\newcommand{\armin}{{\mbox{argmin}}}
\newcommand{\rank}{{\mbox{rank}}}
\newcommand{\<}{\left\langle}
\renewcommand{\>}{\right\rangle}
\newcommand{\lbar}{\left\|}
\newcommand{\rbar}{\right\|}
\newcommand{\eg}{{\emph{e.g.},}}
\newcommand{\ie}{{\emph{i.e.},}}
\newcommand{\wrt}{{\emph{w.r.t.},}}
\newcommand{\etal}{{\emph{et.al.}}}
\renewcommand{\algorithmicrequire}{\textbf{Input:}} 
\renewcommand{\algorithmicensure}{\textbf{Output:}} 
\renewcommand{\a}{{\bm{a}}}
\renewcommand{\b}{{\bm{b}}}
\renewcommand{\d}{{\bm{d}}}
\newcommand{\e}{{\bm{e}}}
\newcommand{\f}{{\bm{f}}}
\newcommand{\g}{{\bm{g}}}
\renewcommand{\o}{{\bm{o}}}
\newcommand{\p}{{\bm{p}}}
\newcommand{\s}{{\bm{s}}}
\renewcommand{\u}{{\bm{u}}}
\renewcommand{\v}{{\bm{v}}}
\newcommand{\w}{{\bm{w}}}
\newcommand{\x}{{\bm{x}}}
\newcommand{\y}{{\bm{y}}}
\newcommand{\z}{{\bm{z}}}
\newcommand{\balpha}{{\bm{\alpha}}}
\newcommand{\bmu}{{\bm{\mu}}}
\newcommand{\bsigma}{{\bm{\sigma}}}
\newcommand{\blambda}{{\bm{\lambda}}}
\newcommand{\bgamma}{{\bm{\gamma}}}

\newcommand{\ba}{{\bm{A}}}
\newcommand{\bb}{{\bm{B}}}
\newcommand{\bc}{{\bm{C}}}
\newcommand{\bd}{{\bm{D}}}
\newcommand{\be}{{\bm{E}}}
\newcommand{\bg}{{\bm{G}}}
\newcommand{\bi}{{\bm{I}}}
\newcommand{\bj}{{\bm{J}}}
\newcommand{\bl}{{\bm{L}}}
\newcommand{\bp}{{\bm{P}}}
\newcommand{\bq}{{\bm{Q}}}
\newcommand{\bs}{{\bm{S}}}
\newcommand{\bu}{{\bm{U}}}
\newcommand{\bv}{{\bm{V}}}
\newcommand{\bw}{{\bm{W}}}
\newcommand{\bx}{{\bm{X}}}
\newcommand{\by}{{\bm{Y}}}
\newcommand{\bz}{{\bm{Z}}}
\newcommand{\bTheta}{{\bm{\Theta}}}
\newcommand{\bSigma}{{\bm{\Sigma}}}

\newcommand{\A}{{\mathcal{A}}}
\newcommand{\B}{\mathcal{B}}
\newcommand{\C}{\mathcal{C}}
\newcommand{\D}{\mathcal{D}}
\newcommand{\F}{\mathcal{F}}
\renewcommand{\H}{\mathcal{H}}
\renewcommand{\L}{\mathcal{L}}
\newcommand{\X}{\mathcal{X}}
\newcommand{\Y}{\mathcal{Y}}

\title{Bringing Giant Neural Networks Down to Earth with Unlabeled Data}
\author{Yehui~Tang, Shan~You, Chang~Xu, Boxin~Shi, \IEEEmembership{Member, IEEE} and Chao~Xu
	\thanks{Yehui Tang and Chao Xu are with the Key Laboratory of Machine Perception (Ministry of Education) and Coopertative Medianet Innovation Center, School of EECS, Peking University, Beijing 100871, P.R. China. E-mail: yhtang@pku.edu.cn, xuchao@cis.pku.edu.cn.}
	\thanks{Shan You is with SenseTime Research. E-mail: youshan@sensetime.com.}
	\thanks{Boxin Shi is with the National Engineering Laboratory for Video Technology, School of EECS, Peking University, Beijing 100871, P.R. China. E-mail: shiboxin@pku.edu.cn.}
	\thanks{Chang Xu is with the UBTech Sydney Artificial Intelligence Centre and the School of Information Technologies in the Faculty of Engineering and Information Technologies at The University of Sydney, J12 Cleveland St, Darlington NSW 2008, Australia. Email: c.xu@sydney.edu.au.}}

\maketitle

\begin{abstract}
Compressing giant neural networks has gained much attention for their extensive applications on edge devices such as cellphones. During the compressing process, one of the most important procedures is to retrain the pre-trained models using the original training dataset. However, due to the consideration of security, privacy or commercial profits, in practice, only a fraction of sample training data are made available, which makes the retraining infeasible. To solve this issue, this paper proposes to resort to unlabeled data in hand that can be cheaper to acquire. Specifically, we exploit the unlabeled data to mimic the classification characteristics of giant networks, so that the original capacity can be preserved nicely. Nevertheless, there exists a dataset bias between the labeled and unlabeled data, which may disturb the training and degrade the performance. We thus fix this bias by an adversarial loss to make an alignment on the distributions of their low-level feature representations. We further provide theoretical discussions about how the unlabeled data help compressed networks to generalize better. Experimental results demonstrate that the unlabeled data can significantly improve the performance of the compressed networks. 
\end{abstract}


\section{Introduction}\label{Introduction}

%
%
%
%
\IEEEPARstart{D}{eep} learning has demonstrated state-of-the-art performance in many tasks, such as object classification~\cite{QiaoZ0WY18,WenHSZCL18}, speech recognition~\cite{NagamineM17,OchiaiWHH17}, image generation~\cite{RezendeMDGW16,MaaloeSSW16} and so on. The major component underlying these successes is the development of sophisticated deep neural networks (DNNs), \eg~AlexNet~\cite{krizhevsky2012imagenet}, VGGNet~\cite{simonyan2014very}, Inception~\cite{szegedy2015going}, and ResNet~\cite{he2016deep}. However, large volume of parameters, huge run-time memory cost and heavy dependence on GPU devices hamper the deployment of these giant DNNs in real-world applications~\cite{wang2017beyond,frankle2018lottery}. For example, ResNet-50~\cite{he2016deep} needs 95MB memory to store parameters, 97MB memory to store feature maps and $3.8 \times 10^9$ times of floating number multiplications to interface with a single image~\cite{wang2017beyond}. It has been well known that there is significant redundancy in a large over-parameterized network, and fewer parameters can express the same amount of information as well~\cite{HanMD15,alvarez2017compression}. It therefore motivates the research on neural network compression.


Basically, network compression methods can be categorized into several aspects, including parameter quantization~\cite{hubara2016binarized, hu2018hashing}, low-rank approximation~\cite{DBLP:conf/nips/DenilSDRF13,DBLP:journals/pami/ZhangZHS16}, knowledge distillation~\cite{hinton2015distilling,wang2018adversarial} and pruning (non-structured pruning~\cite{ReagenGAMRWB18,carreira2018learning,wang2018packing,alvarez2016learning} \& channel pruning~\cite{liu2017learning}). Quantization methods represent weights or activations with low bit integers~\cite{hubara2016binarized}; even binary weights are used~\cite{courbariaux2015binaryconnect,rastegari2016xnor} while low-rank approximation takes advantage of tensor factorization techniques and decomposes one giant filter into multiple smaller components. Knowledge distillation focuses on the training of a compact light network and advocates the soft supervision from pre-trained powerful giant networks~\cite{hinton2015distilling}. As for pruning methods, non-structured pruning~\cite{carreira2018learning} removes unimportant weights, which can get extremely high compression rate without accuracy loss; however, special hardware is needed to accelerate the computation in practice. In contrast, channel-wise pruning~\cite{yu2017nisp,liu2017learning} chooses to remove the whole spatial filters over channels and results in simultaneous reduction on the parameters, memory footprint and computation cost. Note that channel pruning does not destroy the structure of the giant network, so it is compatible with other compression methods and has attracted much attention recently~\cite{DBLP:journals/jzusc/ChengWLHL18}.

Existing neural network compression methods have received impressive performance in experiments,  but they usually need many iterations of retraining to preserve the original accuracies, especially when the compression ratio is fairly high. An immediate question therefore arises: How to retrain the compressed network if the original training dataset is incomplete? Demo models (\eg~well-trained DNNs) are usually released  and ready to import for users. However, due to the consideration of security, privacy or commercial profits, the model providers only supply sample data (sometimes with unknown sources) for verification purpose instead of the complete training set. 
For example, the website of EmotionNet \footnote{\url{https://github.com/co60ca/EmotionNet}} for emotion recognition only provides example images for display, and the original dataset can only be obtained by the approval from administrators with a rigorous agreement. Likewise, powerful CNNs are trained to predict hashtags on massive Instagram images~\cite{mahajan2018exploring}, but the original training dataset is not released as mentioned in the paper except for a few of example images. This situation is more common in medical diagnosis~\cite{djuric2017precision}, drug discovery and  toxicology~\cite{burbidge2001drug}, since their used datasets are usually not completely open-source as discussed in~\cite{cheng2018recent}. In addition, for model compression service on the cloud, uploading the entire training dataset (\eg~several GB file size) to the cloud is unpractical and time-consuming due to the limited accessing speed (\eg~several MB/s), while the trained networks (usually with MB level size) and part of the dataset can be transmitted easily. All these practical senarios imply the necessity of compressing models with no access to  the complete training dataset. However, retraining the compressed models only with the limited labeled data may incur severe over-fitting issue.

In this paper, we propose to bring giant neural networks down to earth with unlabeled data. Instead of struggling to search for original training data, we turn our attention to unlabeled data in hand that can be cheaper to acquire. The output of giant network reflects its classification characteristics and contains the necessary information for its capacity. Thus we regard unlabeled data as a portal to distill its intrinsic information, and concretely, we exploit unlabeled data to mimic the softened output of the giant network, so that its powerful classification ability can be well preserved. However, since unlabeled and labeled data are usually collected in different ways, there is a dataset bias hampering the performance. We fix this issue by make alignment on the distributions of low-level features between unlabeled and labeled data. A confidence coefficient is also introduced to weight the unlabeled data to reduce the disturbance of improper unlabeled examples. Furthermore, we provide theoretical discussions about how the unlabeled data help compressed networks to generalize better. Experimental results on benchmark datasets demonstrate the effectiveness of exploiting unlabeled data to assist the network compression. 

The rest of this paper is organized as follows. Section \ref{sec:rel} reviews the related work for compressing and accelerating networks. Section \ref{sec:cp} makes a brief introduction of channel pruning with scaling factors as preliminaries. Then we formally elaborate in Section \ref{sec:un} how to prune networks with unlabeled data and analyze the proposed method from a theoretical perspective. The experimental results and analysis are presented in Section \ref{sec:ex}, with concluding remarks given in Section \ref{sec:con}.


\section{Related Work}
\label{sec:rel}

For the compression and acceleration of CNNs, the mainstream works are mainly divided into four categories: quantization, sparse or low-rank approximation, knowledge distillation and pruning (non-structured pruning  \& channel pruning). 

\textbf{Quantization.} It aims to reduce the number of bits for representing each weight or activation in the CNNs. For example, Vanhoucke \etal~\cite{vanhoucke2011improving} finds that the 8-bit quantization of weights can induce significant speed-up with almost no drop of accuracy.  Binary weights are even investigated to obtain extremely compressed networks, which constrains weights to only two values (\ie~1 or -1) and most time-consuming multiply-accumulate operations are replaced by simple accumulations ~\cite{courbariaux2015binaryconnect,rastegari2016xnor}. However, binarizing very large networks (\eg~GoogleNet) will incur large accuracy loss. To improve the performance of quantized networks, Li \etal~\cite{li2016ternary} proposes Ternary Weight Network (TWN) constraining weights to ternary values (\ie~-1,0,1). Zhu \etal~\cite{zhu2016trained} further develops it by learning both ternary values and assignment during training. The proposed Trained Ternary Quantization (TTQ) can be trained from scratch as easy as a normal full-precision model. 


\textbf{Low-rank approximation. } Since convolutional filters can be seen as 4D tensors, based on low-rank assumption, they can be decomposed to multiple components with fewer parameters. Both storage and computation cost can be reduced in the way. For example, SVD method has been studied widely~\cite{DBLP:conf/nips/DenilSDRF13,DBLP:journals/pami/ZhangZHS16} to decompose a tensor into two-layer compact convolutional filters. Those components with large sizes 
may be still computationally time-consuming, which can be further  decomposed\cite{DBLP:journals/pami/ZhangZHS16}. Thus an original redundant filter is replaced by multiple compact filters.

\textbf{Knowledge distillation.} By distilling the knowledge from the pre-trained giant networks, the performance of the target light and small networks can be boosted~\cite{hinton2015distilling,yim2017gift}. Hinton \etal~\cite{hinton2015distilling} proposes to mimic the informative softened outputs of the teacher network. In addition to the output level, intermediate representations of the giant network can be transferred as hints to assist training. In~\cite{zagoruyko2016paying}, the attention maps via activations or gradients are also used for mimicking. Besides, You \etal~\cite{you2017learning} proposes to combine multiple teacher networks. The relative dissimilarity between different examples serves as guidance and a voting strategy is used to unify dissimilarity information provided by various teacher networks. There are also some works~\cite{li2017learning,su2016adapting} extending knowledge distillation to more applications such as multi-task learning~\cite{li2017learning}. 

\textbf{Non-structured pruning.} To prune the redundant parameters, an intuitive method is to remove each weight with small magnitude and get a more sparse network. Han \etal~\cite{han2015learning} proposed to apply $l_1$ or $l_2$ regularization to make weights sparse and prune tiny weights in an iterative way. The pruned network can be further compressed  with quantization and Huffman encoding, resulting in $35\times$ compression rate on AlexNet without sacrifice for accuracy ~\cite{han2015deep}. To avoid accuracy drop incurred by incorrect pruning weights, splicing operation~\cite{NIPS2016_6165} was introduced to recover the mistakenly removed connections. Pruning and splicing operations constitute the dynamic network surgery framework and obtain more sparse networks with fewer training epochs. However, although high compression and acceleration rates are obtained theoretically,  the hardware is needed to designed specially for realizing practical speed-up. Compared to the fine-grained pruning methods, group-wise pruning methods~\cite{DBLP:conf/cvpr/LebedevL16} are more common in practice. Nevertheless, structures of the original networks are destroyed as a result and real inference speed-up also depends on dedicated libraries badly~\cite{li2016pruning,gao2018dynamic}. 

\textbf{Channel pruning.} Channel pruning methods aim to directly remove the redundant channels without destroying the structure of original networks. After pruning a whole filter of a layer, the channels of the corresponding feature maps are also pruned. Parameters, computation cost and memory footprint are reduced simultaneously. There are mainly two strategies for channel pruning. The first one~\cite{he2017channel,luo2018thinet} seeks to identify the important channels layer-by-layer by minimizing the gap of feature maps between the pruned network and the original pre-trained network. Though the layer-wise reconstruction can preserve the information of each individual layer well, it ignores the global information of the entire network which  obstructs  deciding the width of each layers automatically and achieving high compression rate. The more prevalent strategy is to select important channels for various layers simultaneously and  train the whole networks with sparsity regularization~\cite{wen2016learning,liu2017learning,alvarez2016learning,molchanov2019importance}. Slimming method~\cite{liu2017learning} uses the scaling factors of batch normalization layers~\cite{ioffe2015batch} to measure the importance for each channel. During training, the sparse constraint is imposed on the scaling factors and then channels with tiny scaling factors are pruned. The pruned networks are then fine-tuned with common cross-entropy loss to recover the performance. Recently, it was  empirically found that when the training data are available, training the pruned network from scratch might achieve higher performance~\cite{liu2018rethinking}. However, for situation that the entire training dataset is not accessible, the pre-trained weights in the giant network are vital to get a pruned network with high performance. 

\section{Channel-wise Pruning under Sparse Regularization }
\label{sec:cp}

A number of well-trained DNNs can be easily obtained from the Internet and tailored for various tasks. Most of the time these downloaded networks are too cumbersome to be applied directly in practical tasks especially for those on edge mobile devices. So some questions arise immediately: How many parameters would be sufficient for DNNs to reach decent performance? How much computational budget can be offered by our computing platforms? Answers to these questions are not unique, depending on different real-world applications. It is therefore impossible to request model providers to release well-trained models of various sizes from a few hundred KB to several hundred MB to meet all users' demand. A practical solution is to compress the released giant models to an appropriate size that can meet different requirements. In the sequel, we will revisit how the giant neural networks can be compressed to a specific size using channel pruning techniques. Moreover, we also illustrate that why channel pruning would be degraded when the labeled data are quite limited.

Suppose a dataset $\D = \{(\x_i,\y_i)\}_{i=1}^N$ \footnote{In our problem, the dataset capacity $N$ for labeled data are usually small.} containing $N$ examples $\x_i\in\X \subset \mathbb{R}^d$ and corresponding labels $\y_i\in\Y\subset \mathbb{R}^K$ are released with a well-trained DNN, where $\X$ and $\Y$ are raw feature space and label space, respectively. Denote the released well-trained neural network of $L$ layers as a function $\tilde{f} \in\F$ , where $\mathcal{F}$ denotes the hypothesis space of DNNs. A (channel-wise) sparse network is usually trained by minimizing the objective function defined on these labeled data, \ie
\begin{equation}
\label{sparseori}
\mathcal{L}_l (f) = \frac{1}{N} \sum_{(\x_i,\y_i)\in\D} \L_{o} (f(\x_i,W), \y_i) + \lambda \cdot\mathcal R (W)
\end{equation}
where  $\L_{o}$ is a supervision loss (\eg~cross entropy for image classification) to guarantee the network performance and $\mathcal R (\cdot)$ is a  sparse regularizer imposed on the weights. $\lambda$ balances the classification accuracy and the sparsity of weights, and a larger $\lambda$ induces a sparser network. To get a network with channel-wise sparsity that can be inferred fast without specific hardware/software support, $\mathcal R (\cdot)$ can serve as a group sparsity over weights directly~\cite{wen2016learning}. However, implementing the group sparsity on the massive weights also increases the difficulty in optimization\cite{liu2017learning},  incurring slow convergence and unsatisfied results. Hence the scalar factor $\bgamma=\{\gamma^{(j)}\}_{j=1}^m$ is introduced as surrogates to control the channel-wise sparsity.~
Denoting $\z_i$ as  the $l$-th layer in the feature map of the network $f$ for example $\x_i\in \X$, the feature map $\hat{\z}_i$ after scaling is obtained as:
\begin{equation}
\hat{\z}_i^{(j)}= \gamma^{(j)} \z_i^{(j)} ,
\end{equation}
where $\hat{\z}_i^{(j)}$ and ${\z}_i^{(j)}$ are the $j$-th channel of $\hat {\z}_i$ and ${\z}_i$, respectively. The scaling factor $\bgamma$ control the information flow and a small $\gamma^{(j)}$ blocks the information of the corresponding channels. 
In practice, the trained parameters in Batch Normalization~(BN) layers can work as the scaling factors\cite{liu2017learning, ye2018rethinking} and for those networks without BN layers, it is also easy to insert the extra scaling factors to scale feature maps in the training phase. 

When training the network,  sparsity regularization is imposed on scaling factors $\bgamma$'s, and the objective function evolves into:
\begin{equation}
\label{sparse}
\mathcal{L}_l (f)= \frac{1}{N} \sum_{(\x_i,\y_i)\in\D} \L_{o} (f(\x_i,W,\Gamma), \y_i) + \lambda \cdot \norm{\Gamma}_1,
\end{equation}
where $\norm{\cdot}_1$ is $\ell_1$ norm to encourage only part of channels are selected to establish the network, and $\Gamma$ is a vector composed of all scaling factors over the whole network.
A network with spare scaling factors is obtained via Eq.~\eqref{sparse}, and then the  network is pruned based on the values of scaling factors. A global threshold across layers is set according to the percentage of channels users plan to keep. For example, if users want to prune 60\% channels of the network, the smaller 60\% elements in $\Gamma$ are removed with their corresponding channels. The structure of the network will  ta automatically decided according to the threshold, and a pruned network is obtained as a result. After pruning, the network usually has limited classification accuracy. Fine-tuning the pruned network is therefore essential to restore the accuracy. Typically, when we fine-tune the pruned network, the objective is only the supervision loss (\ie~the first term in Eq. \eqref{sparse}).

However, only with the extremely limited data released, Eq. \eqref{sparse} (with or without sparsity term $\norm{\Gamma}_1$) cannot be well optimized. Serious overfitting will occur and the accuracy on the test set will drop rapidly, incurring poor performance of the pruned network. 

\section{Pruning networks with unlabeled data}
\label{sec:un}
Few labeled data limit the performance of channel-pruned networks. Instead of accepting the poor performance of the pruned network with only limited sample data or struggling to search for original training data, we turn attention to cheaper unlabeled data in hand. In this section, we will present a solution to bring giant networks down to earth with channels pruned, in which we will investigate the potential benefits from unlabeled data. 

\subsection{Exploiting Unlabeled Data by Mimicking the Giant Network}
Unlabeled data are much easier to collect. For example, one can easily find a large number of natural images from the Internet to assist compression of giant networks trained on large natural images set. The unlabeled data collected by users may be different from the original data used for the well-trained giant network, but still can provide helpful information for the compression task. 



Suppose the collected unlabeled dataset $\D^u = \{\x^u_i\}$ contains $N^u$ examples $\x^u_i\in\X^u\subset \mathbb{R}^d$ \footnote{Here we assume the labeled data and unlabeled data have the same dimension $d$, which can be easily implemented by image resizing or cropping.}. In the sequel, we distinguish the labeled and unlabeled examples with notations $(\x^l,\y^l)$ and $\x^u$, respectively. 

Similarly, given $\x_i^u \in \mathcal{X}^u$, the \textit{softened} output of the network $f$ and the released well-trained network $\tilde{f}$ after softmax function are represented as $\p_i^u$ and $\tilde{\p}_{i}^{u}$, respectively, 
\begin{align}
\label{soft}
\p_i^u= \frac{\exp( f(\x_i^u)/ \tau )}{\norm{\exp(f(\x_i^u)/ \tau)}_1}, ~
\tilde{\p}_{i}^{u}= \frac{\exp( \tilde{f}(\x_i^u)/ \tau )}{\norm{\exp(\tilde{f}(\x_i^u)/ \tau)}_1},
\end{align}
where $\tau$ is a temperature parameter~\cite{hinton2015distilling} to control the smoothness, so that a higher value of $\tau$ produces a softer probability distribution over classes. The softened output of the giant network reveals clues about its classification characteristics as well as the similarity among classes. Thus we can use unlabeled data to mimic its classification performance and distill its knowledge into the target sparse network by minimizing the cross-entropy $\mathcal{H}(\p_i^u,\tilde{\p}_{i}^{u})$ between softened outputs of labeled and unlabeled data.

\textbf{Confidence on the unlabeled data.} However, the extra collected unlabeled data may have different categories compared with the labeled data for training the giant network, and many examples may  not be well understood even by the giant network. Directing imposing the standard distillation loss $\mathcal{H}(\p_i^u,\tilde{\p}_{i}^{u})$ will also enforce the spare network to mimick the outputs of the giant network on those aberrant data, which  may disturb the learning process on labeled data and incur performance degradation. Thus we propose to  treat   the unlabeled examples  differently by weighting them with the confidence of the giant network $\tilde{f}$. The confidence can be reflected by their corresponding outputs, and for  the network with a high confidence on example $\x_i^u \in \mathcal{X}^u$, one element in its output vector $\tilde{f}(\x_i^u)$  will be far larger than other elements.  Thus the maximum of the  output vector normalized by softmax function can be used as the confidence.  However,  the original softmax function (\ie~with $\tau=1$) is so sharp that the maximum in $\tilde{f}(\x_i^u)$ would be very close to $1$ in most cases. Temperature $\tau$ is used to soften the output vector so that the maximum in $\tilde{f}(\x_i^u)$  would be more sensitive when the confidence changes. The weight $C_i^u$ for the example $\x_i^u \in \mathcal{X}^u$ is defined:
\begin{equation}
\label{conf}
C_i^u=\max \frac{\exp( \tilde{f}(\x_i)/ \tau )}{\norm{\exp(\tilde{f}(\x_i)/ \tau)}_1},
\end{equation}
where $\max(\cdot)$ is the maximum value of a vector. In this way, we encourage the sparse network $f$ to have similar softened outputs with those of the giant network $\tilde{f}$ on the examples with high confidence. Then the objective of the sparse retraining on unlabeled data (\textit{distillation loss}) can be written like Eq. \eqref{sparse} as
\begin{equation}\label{unlabeled}
\mathcal{L}_m (f)=\frac{1}{N^u}\sum_{\x_i^u\in\D^u} C_i^u \cdot \mathcal{H}(\p_i^u,\tilde{\p}_{i}^{u})+\lambda \cdot \norm{\Gamma}_1,
\end{equation}  
where $\mathcal{H}(\cdot,\cdot)$ is the cross-entropy. Considering both labeled and unlabeled data, the objective function for training network $f$ is 
\begin{equation}
\label{losssum}
\begin{split}
\mathcal{L}(f)=& \mathcal{L}_l (f)+ \alpha \cdot \mathcal{L}_{m}(f), \\
\end{split}
\end{equation}
with a constant weight $\alpha\geq0$. In this way, the unlabeled data can help the labeled data by further supplying more information about the giant network. The confidences automatically weight different unlabeled data and discard those improper and noisy data.  
Note that the labeled data can  also be used to mimic the output of the pre-trained giant network\footnote{Then the first term in Eq.\ref{unlabeled} can be replaced by $\frac{1}{N^u}\sum_{ \x_i^u\in\D^u} C_i^u \mathcal{H}(\p_i^u,\tilde{\p}_{i}^{u})+\frac{1}{N^l}\sum_{\x_i^l\in\D^l} \mathcal{H}(\p_i^l,\tilde{\p}_{i}^{l})$, where $\p_i^l= \frac{\exp( f(\x_i^l)/ \tau )}{\norm{\exp(f(\x_i^l)/ \tau)}_1}, 
	\tilde{\p}_{i}^{l}= \frac{\exp( \tilde{f}(\x_i^l)/ \tau )}{\norm{\exp(\tilde{f}(\x_i^l)/ \tau)}_1}$.}. However, the effect of distillation on labeled data is subtle due to their limited quantity, which is further verified in the experiments. 

\subsection{Fixing the Dataset Bias between Unlabeled and Labeled Data}\label{sec:ad}

In Eq. \eqref{unlabeled} above, the unlabeled data are usually utilized to make a consistency of the probabilistic output between the sparse network and the original giant network. However, in practice, the collected unlabeled data by users are usually different from the original labeled data. And there exists a dataset bias (or domain shift)~\cite{quinonero2008covariate} between the unlabeled data and labeled data. As a result, the consistency in unlabeled data may not hold on labeled data. Since both the pruned network and the giant network are designed for labeled data, the classification performance of the pruned network would be degraded due to this dataset bias. 

To make the massive unlabeled data act as the supplement of limited labeled data well, the representations of unlabeled data are expected to be consistent with those of labeled data, which can be fulfilled by make an alignment on the representations from two datasets.  We use a few low-level layers   of the network $f$ as an aligner $f_1$  and denote the remaining part as $f_2$. Both labeled and unlabeled data are first sent to $f_1$ to produce representations with same distributions and then based on these coincident representation, $f_2$ outputs the classification results, \ie~$f(\cdot)=f_2(f_1(\cdot))$.

Generally, the gap between two distributions can  be measured by multiple measures such as maximum mean discrepancy (MMD)~\cite{long2016unsupervised} and various divergence metrics~\cite{goldberger2003efficient}. Following the recent success of adversarial training methods to reduce the gap between different  distributions in many tasks, such as image style transfer~\cite{chang2018pairedcyclegan}, image super-resolution~\cite{ledig2017photo} and domain adaptation~\cite{tzeng2017adversarial}, we minimize the discrepancy between the unlabeled and labeled representation distributions by introducing a discriminator~\cite{goodfellow2014generative}, which distinguishes the distributions from different datasets and is co-trained with a generator in an adversarial learning manner.

The adversarial training~\cite{goodfellow2014generative} can be regarded as a two-player minimax game.  Given examples $\x_i^l \in \mathcal{X}^l, \x_i^u \in \mathcal{X}^u$, the discriminator $D$ is to make binary predictions about whether their low-level feature representations $f_1(\x_i^l), f_1(\x_i^u)$ are from unlabeled dataset or not. In this case, the aligner $f_1$ plays the role as the generator, which tries to confuse the two feature maps. The game can be modelled with a value function $V(f_1,D)$:
\begin{multline}\label{value}
\min_{f_1} \max_D V(f_1,D) = \Exp_{\x^l \sim p^l(\x^l)} \qileft\log (D(f_1(\x^l)))\qiright \\
+ \Exp_{\x^u \sim p^u(\x^u)} \qileft \log(1- D(f_1(\x^u)))\qiright.
\end{multline}
Eq. \eqref{value} is usually solved by alternatively optimizing the $D$ and $f_1$, whose loss functions are 
\begin{align}
\L_D(D) &= - \hat{V}(f_1,D)~\mbox{with fixed}~f_1,\label{loss_D}\\ 
\L_{f_1}(f_1) &= \hat{V}(f_1,D)~\mbox{with fixed}~D,
\end{align}
where $\hat{V}$ is the empirical loss of the value function $V$, \ie~
\begin{multline}\label{v}
\small
\hat{V}= \frac{1}{N^l}\sum_{\x_i^l\in\D^l}\log (D(f_1(\x_i^l))) \\
+\frac{1}{N^u}\sum_{\x_i^u\in\D^u} \log(1- D(f_1(\x_i^u))).
\end{multline}

The optimization of $\L_D(D)$ and $\L_{f_1}(f_1)$ is iterative and at last the distance of low-level features from labeled data and unlabeled data will be minimized. Thus the aligner $f_1$ can produce accordant representations of labeled and unlabeled data for the main body $f_2$.  Note that our scenario is different from the typical unsupervised domain adaptation task~\cite{tzeng2017adversarial,ganin2014unsupervised}, which adapts a network trained with massive labeled data  to achieve high performance in an unlabeled domain by aligning the features in late layers, while neglecting the performance degradation on the original domain.  As the unlabeled data here are only used to  supply the labeled data and  assist compressing networks, the  representation alignment is conducted on the low-level layers to adapt the unlabled data to be consistent to labeled data and  compatible with the pre-trained network. This strategy adequately  utilizes the information of the unlabeled data while reduce the  disturbance  to the normal training from  the   alignment process, and hence we can obtain a well-performanced pruned network.



In practice, the aligner $f_1$ for feature aligning and the main body $f_2$ for outputting prediction results  can be trained end-to-end by augmenting the original objective Eq. \eqref{losssum} with the \textit{adversarial loss} $\L_{f_1}(f_1)$, \ie~
\begin{equation}\label{key}
\L(f) = \mathcal{L}_l (f)+ \alpha \cdot \mathcal{L}_{m}(f) + \beta \cdot \L_{f_1}(f_1),
\end{equation}
where $\beta\geq0$ is the weight coefficient.  In Eq.~\eqref{key}, the   unlabeled data are first adapted to be compatible with the network by representation alignment, and then different examples are weighted by 
the example-wise  confidence weight $C_i^u$ (Eq.~\eqref{conf}). Hence the information in the massive unlabeled data are adequately explored and improper examples are also filtered. As a consequence, the sparse network $f$ can well receive the help from unlabeled data for mimicking the giant network, but with subtle influence by the dataset bias. 

\subsection{Theoretical Discussions}
Now we attempt to investigate how the unlabeled data help the pruned network to generalize better than that with only a few labeled data. For simplicity of theoretical discussions, the training of entire system can be divided into two steps sequentially. In the first step, we adversarially train the aligner $f_1$ and the discriminator $D$; then in the second step, the main body of network $f_2$ is trained to mimic the output of the original giant network $\tilde{f}$. 

First, using the unlabeled and labeled data at low-level layers, we train the network via Eq. \eqref{value}. Then in theory, we can make their distributions identical on feature representations, via the following Theorem \ref{th:gan}~\cite{goodfellow2014generative}.
\begin{theorem}[Feature alignment]\label{th:gan}
	With $f_1$ being fixed, the optimal discriminator $D$ is $D^*(\x) = p^l(\x)/(p^l(\x)+p^u(\x))$. Then the global optimality is achieved if and only if $p^l=p^u$.
\end{theorem}

As a result, we can align the distribution of unlabeled data's low-level features with that of labeled data. Since $p^l=p^u$, the input of the main body $f_2$ would have no dataset bias in theory. Then $f_2$ can be trained with the loss of Eq.~\eqref{losssum}, which can be cast into the framework of empirical risk minimization (ERM) with regularization. To facilitate the analysis on the unlabeled data's effect, we leave out the regularization term. 
Then we investigate the generalization ability of the learned $f_2$ by checking its generalization error bound, which is related to its population risk and empirical risk defined as
\begin{align}
\small
R(f_2)&= \Exp_{(\x,\y) \sim \mathcal{Q}}\qileft \ell(f_2(\x),\y)\qiright, \\
\hat{R}(f_2)&= \frac{1}{N} \sum_{(\x_i,\y_i) \in \D} \ell(f_2(\x_i),\y_i),
\end{align}
where $\mathcal{Q}\in\X\times\Y$ is the ground-truth distribution of $(\x,\y)$\footnote{Here we do not distinguish the hard label vector and softened output vector in Eq. \eqref{losssum}, and regard both as the target space $\Y$ for simplicity. }. Usually, there exists a gap between the population risk $R(f)$ and empirical risk $\hat{R}(f)$. A desired model should have small gap. Via MaDiarmid's inequality, the gap can be bound by Theorem \ref{gap}~\cite{koltchinskii2002empirical}.
\begin{theorem}[Generalization error bound]\label{gap}
	Given a fixed $\rho >0$, for any $\delta>0$, with probability at least $1- \delta$, for all $f \in F$
	\begin{equation}
	R(f_2) \le \hat{R}(f_2)+\frac{2K^2}{\rho N} R'_{N}(F)+ \sqrt{\frac{\ln{\frac{1}{\delta}}}{2N}},
	\end{equation}	
	where $K$ is the number of classes, and $R'_{N}(F)$ is the Rademacher complexity. 
\end{theorem}
In Theorem \ref{gap}, the third term shows that a large dataset capacity $N$ induces a tight bound. In this way, with the unlabeled data involved, we boost the generalization ability of $f_2$ by increasing the training examples, which have identical distributions after the feature alignment. The second term refers to the Rademacher complexity as follows.
\begin{Definition} \label{def}
	Given rademacher variables $\sigma_i$ (independent uniform random variables in \{-1,+1\}) and $\x_i \in \X$, the Rademacher complexity of hypothesis $F \ni f$ is defined as 
	\begin{equation}
	\small
	R'_{N}(F)=\Exp_{\mathcal{X},\sigma_i}\qiLeft \sup_{k,f^k} \sum_{i=1}^{N} \sigma_i f^k(\x_i)\qiRight,
	\end{equation}
	where $f^k(\x_i)$ is the $k$-th element in the output vector $f(\x_i)$.
\end{Definition}
$R'_{N}(F)$ is directly related with the complexity of the hypothesis and the generalization gap as well. Thus to make the gap tighter, we can train the network $f_2$ by minimizing $R'_{N}(F)/N$. However, its computation is very hard. In practice, we usually use its upper bound or estimation~\cite{kawaguchi2017generalization} for each minibatch, \eg~
\begin{equation}
\label{re2}
R_c(f)=\frac{1}{N'}\max_k \sum_{i=1}^{N'} |f^k(\x_i)|,
\end{equation}
where $N'$ is the number of both labeled and unlabeled samples in a minibatch. Then term $R_c(f)$ can thus serve as a regularization term (called \textit{Rademacher loss}) to control the generalization ability during the training of the network; the loss function is 
\begin{equation}\label{losssum4}
\L_{all}(f) = \mathcal{L}_l (f)+ \alpha \cdot \mathcal{L}_{m}(f) + \beta \cdot  \L_{f_1}(f_1) +\eta \cdot  R_c(f),
\end{equation}  
where $\eta\geq0$ is a constant parameter, and $R_c(f)$ is calculated per minibatch. Although unlabeled data are much cheaper than labeled data, they are not free and large disks are needed to store the collected unlabeled data. Over-fitting usually occurs when tailoring or retraining the giant network with insufficient data, and thus in this case promoting the generalization with $R_c(f)$ will work. Our proposed method is summarized in Algorithm \ref{alg1}. Similarly, it also contains three steps, and unlabeled data play an important part in the sparse retraining and fine-tuning to assist the labeled data.  

\begin{algorithm}[t]
	\caption{Pruning with Unlabeled Data (PUD)}
	\label{alg1}
	\begin{algorithmic}[1]
		\REQUIRE Pretrained network $\tilde{f}$, released labeled dataset $\D^l$ and collected unlabeled dataset $\D^u$
		\STATE	Initialize the sparse network $f$ with $\tilde{f}$.
		\REPEAT
		\STATE Randomly select $\x_i^l\in \D^l$ and $\x_i^u\in \D^u$ as a minibatch.
		\STATE Forward the pre-trained giant network:\\
		$\{\tilde{\p}^l_i, \tilde{\p}^u_i, C_i^u\} \leftarrow \{\tilde{f}(\x_i^l),\tilde{f}(\x_i^u)\}$.
		\STATE Forward the network: \\
		$\{\p_i^l, \p_i^u, f_1(\x_i^l), f_1(\x_i^u)\} \leftarrow \{f(\x_i^l),f(\x_i^u)\}$.
		\STATE Calculate the loss of discriminator $D$ with Eq. \eqref{loss_D} and update the parameters of $D$.
		\STATE Calculate the loss of network $f$ with Eq.~\eqref{losssum4}.
		\STATE Update the parameters of network $f$.
		\UNTIL{convergence}
		\STATE Prune channels with small scaling factors in network $f$.
		\STATE Fine-tune the pruned network. 
		\ENSURE A pruned network ready to deploy. 
	\end{algorithmic}
\end{algorithm}

\section{Experiments}
\label{sec:ex} 
In this section, we empirically verify the effectiveness of the proposed method pruning networks with unlabeled data (PUD), which is compared with several state-of-the-art methods including SSL~\cite{wen2016learning}, PFEC~\cite{li2016pruning}, Slimming~\cite{liu2017learning} and ISTA-Pruning~\cite{ye2018rethinking}. SSL~\cite{wen2016learning} and PFEC~\cite{li2016pruning} identify unimportant filters by directly checking the network weights, while Slimming~\cite{liu2017learning} prunes networks according to the scaling factors in BN layers. More recently, ISTA-Pruning~\cite{ye2018rethinking} adopts ISTA~\cite{beck2009fast} to solve the spare-constrained optimization and achieves good performance.  A Vanilla Pruning method is also conducted  as a baseline, which just removes the small scaling factors of the giant networks and then fine-tunes the pruned networks with the labeled data. In addition, we train the pruned networks from scratch by randomly initializing their parameters, which is denoted as `Scratch' method in our experiment. As these existing compression methods have be well optimized to achieve state-of-the-art performance on benchmark datasets including CIFAR-10~\cite{krizhevsky2009learning} and large-scale Imagenet (ILSVRC2012)~\cite{ILSVRC15}, for fair comparison, we conduct experiments on the them together with the prevalent VGGNet and ResNet following~\cite{liu2017learning}. As for the assistant unlabeled data, we adopt the STL-10 dataset~\cite{CoatesNL11} and COCO dataset~\cite{LinMBHPRDZ14}, respectively.

\begin{table*}[t]
	\caption{Classification accuracy of the pruned VGGNet on CIFAR-10 dataset with the unlabeled STL-10 dataset. All methods achieve approximately $11\times$ compression rate (1.8M Params) and $2.5\times$  acceleration rate (160M FLOPs).} 
	\label{cfvgg}
	\centering
	\small		
	\begin{tabular}{l||c|c|c|c|c|c|c|c}\hline
		\multirow{5}{*}{VGGNet}&\multirow{2}{*}{$N^l$}&\multirow{2}{*}{Scratch}&\multirow{2}{*}{Vanilla Pruning}&SSL&PFEC&Slimming&ISTA-Pruning&\multirow{2}{*}{PUD (Ours)} \\ 
		&&&&\cite{wen2016learning}&\cite{li2016pruning}&\cite{liu2017learning} &\cite{ye2018rethinking}& \\  \cline{2-9}
		&100&41.47&59.28&61.01&61.04&62.49&63.32 &\textbf{75.04}\\ 
		&500&56.97&78.31&79.42&80.21&84.52&85.93&\textbf{88.51}   \\  
		&1K&69.86&82.56&83.23&84.69&87.23&88.05&\textbf{91.14}   \\   
		\hline
	\end{tabular}
\end{table*}

\begin{table*}[t]
	\caption{Classification accuracy of the pruned ResNet-56 on CIFAR-10 dataset  with the unlabeled STL-10 dataset. All methods achieve approximately $2.0\times$ compression rate (0.3M Params)  and $2.5\times$ acceleration rate (35M FLOPs).}
	\label{cfres}
	\centering
	\small
	\begin{tabular}{l||c|c|c|c|c|c|c|c}\hline
		\multirow{5}{*}{ResNet}&\multirow{2}{*}{$N^l$}&\multirow{2}{*}{Scratch}&\multirow{2}{*}{Vanilla Pruning}&SSL&PFEC&Slimming&ISTA-Pruning&\multirow{2}{*}{PUD (Ours)} \\ 
		&&&&\cite{wen2016learning}&\cite{li2016pruning}&\cite{liu2017learning} &\cite{ye2018rethinking} &\\  \cline{2-9}
		&200&42.32&51.49&52.65&54.01&55.48&55.76&\textbf{62.03} \\ 
		&500&52.78&65.78&66.21&66.43&67.02&68.61&\textbf{73.29}   \\  
		&1K&65.24&77.19 &78.38&78.49&79.22&79.82&\textbf{82.42}   \\  
		\hline  	
	\end{tabular}	
\end{table*}

\subsection{Experiments on CIFAR-10 Dataset}
\begin{figure}[t]
	\centering
	\subfigure[CIFAR-10 dataset.]
	{\includegraphics[width=0.45\columnwidth]{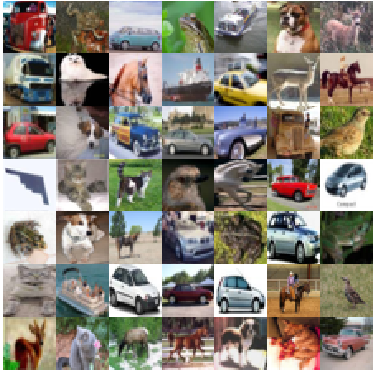}}
	~	\subfigure[STL-10 dataset.]
	{\includegraphics[width=0.45\columnwidth]{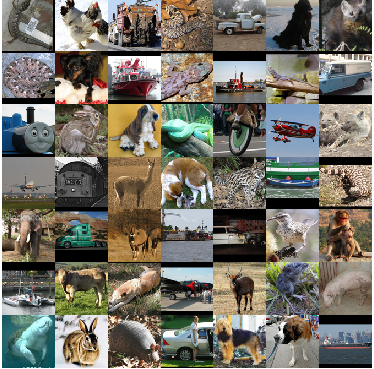}}
	\caption{Sample images in the labeled CIFAR-10 dataset~\cite{krizhevsky2009learning} and the unlabeled STL-10 dataset~\cite{CoatesNL11}.}
	\label{pics:cfstl}
	\vskip -0.2in
\end{figure}

\textbf{Dataset.} The CIFAR-10 dataset~\cite{krizhevsky2009learning} is composed of 60,000 $32 \times 32$ color images from ten categories, 50,000 for training and 10,000 for testing. In our setting, only a small fraction of images are randomly selected as labeled data. The standard data argumentation~\cite{he2016deep} is adopted, including padding (with size 4), random cropping and horizontal flipping. As for the unlabeled data, we choose STL-10 dataset~\cite{CoatesNL11}, which is also an image recognition dataset containing a large number of $96 \times 96$ (labeled and unlabeled) RGB images. Actually, STL-10 dataset has similar categories with CIFAR-10 dataset, however, their collection approaches are different. Some example images of the two datasets are shown in Figure \ref{pics:cfstl}.
In our experiment, we randomly sample 5000 images from the unlabeled part of the STL-10 dataset to assist the compression. All unlabeled images are then rescaled into the same size $32 \times 32$.   

\textbf{Networks.} We experiment with VGGNet~\cite{simonyan2014very} and ResNet-56~\cite{he2016deep}, which are deep and powerful baseline networks broadly used in many tasks, such as image recognition, objection detection and video action analysis. The original VGGNet is designed for ImageNet dataset, thus we tailor its structure slightly to fit CIFAR-10 dataset following~\cite{liu2017learning}.  The features extracted by the convolution layers are pooled by a $2\times2$ pooling layer and then directly sent to a fully-connected layer to obtain predictions. The 56-layer ResNet is stacked by bottleneck blocks with pre-activation structure\cite{he2016identity}. We train the VGGNet and ResNet from scratch in CIFAR-10 dataset as the giant pre-trained networks. For the adversarial loss in Section \ref{sec:ad}, we adopt the second pooling layer in VGGNet and the first block in ResNet as low-level feature layers, and the discriminator $D$ is a simple 3-layer CNN. Feature maps are first delivered into  two convolution layers followed by ReLU nonlinear operation, then forwarded to an average-pooling layer and a  fully-connected layer to predict whether the image comes from labeled dataset or unlabeled dataset. The number of output channels of the first convolution layer is equal to the number of its input channels while the second convolution layer has double channels.

\textbf{Training.} For sparse retraining with Eq. \eqref{losssum4}, roughly equal iterations (15K$\sim$20K) are used. We experimentally find that this training iteration number suffices for both comparison methods (using only labeled data) and our method (using both labeled and unlabeled data). As for fine-tuning the pruned network, we use half of the iterations, \ie~7.5K$\sim$10K. For VGGNet, the initial learning rate is set to 0.003 in sparse retraining and 0.001 in fine-tuning, and for ResNet, it is set to 0.02 and 0.005, respectively. Learning rate drops by 0.1 at $1/2$ and $3/4$ of the maximum iterations for training with only labeled data. For training with additional unlabeled data, it drops by 0.3 at 40\%, 70\% and 90\% of the maximum iterations. We empirically find that the two learning rate schemes fit their own setting well.  For VGGNet, we select $\lambda$ in the interval  $[0.0010,00015]$ with step 0.0001 to control the sparsity of the network via the term $\norm{\bgamma}_1$, and for ResNet we select in the set  $\{0.001,0.002,0.003\}$. When calculating the loss of discriminator, the labeled data are weighted by a coefficient equal to $N^u/N^l$ for balance. The weight $\alpha$ and the temperature parameter $\tau$ are set to 0.7 and 3, respectively. The weight of adversarial loss $\beta$ is set to $10^{-6}$, while the weight of Rademacher loss $\eta$ is select from $\{0.01,0.001\}$. Parameters are determined with cross-validation.

\begin{figure*}[t]
	\centering
	
	\includegraphics[width=1.99\columnwidth]{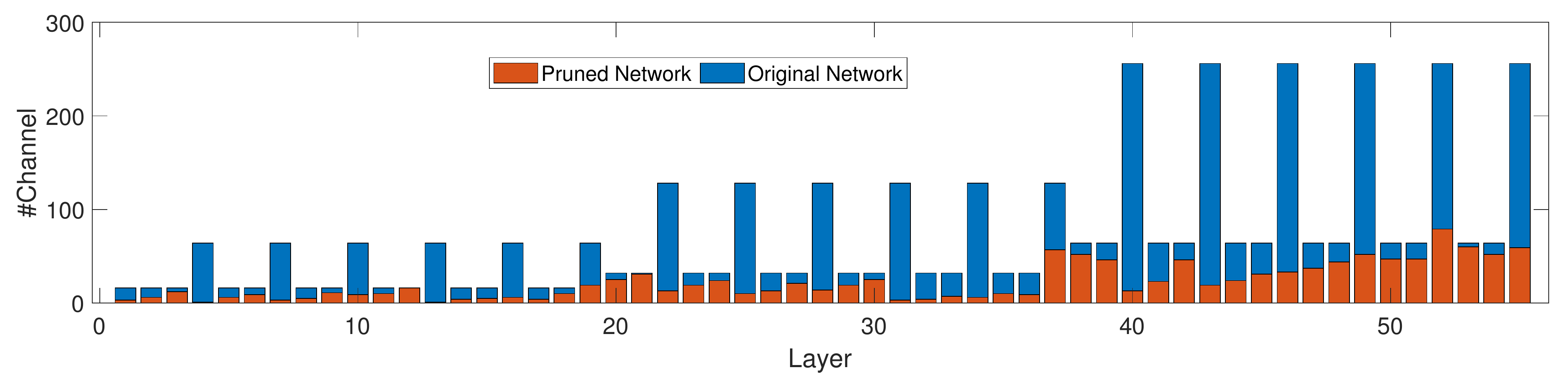}	
	\centering
	\caption{Detailed struture of ResNet-56 on CIFAR-10 dataset by the proposed PUD method. The blue bar denotes the number of channels of the giant network while the red bar denotes that of the pruned network.}
	\label{cifresstr}
	
\end{figure*}

\textbf{Results.} The classification accuracy of the pruned networks on CIFAR-10 dataset assisted by STL-10 dataset is presented in Table \ref{cfvgg} and Table \ref{cfres} for VGGNet and ResNet, respectively. The pre-trained VGGNet (ResNet) achieves 93.78\% (93.96\%) accuracy with 20.1M (0.59M) parameters and 398.6M (88.3M) float-point-operations (FLOPs). For fairness of comparison, all methods prune 70\% channels of the pre-trained models assisted with 5K unlabeled images from STL-10 dataset, and obtain pruned networks with about 1.8M (0.3M) parameters and 159M (35M) FLOPs.  \footnote{The actual compression rate and acceleration rate are related to the percentage of channels pruned in each layer and may vary in a small range.} 

From Table \ref{cfvgg} and Table \ref{cfres}, we can see that with various numbers of labeled images $N^l$, the proposed PUD method significantly outperforms the comparison methods in all cases. This indicates the effectiveness and superiority of  exploiting unlabeled data even when the distributions of labeled and unlabeled images are not exactly identical. When the labeled data are not sufficient, the comparison methods tend to be trapped in serious over-fitting problem. For example, with 1K labeled data, the state-of-the-art Slimming method only achieves 87.23\% accuracy, with a large accuracy drop (6.55\%) from the pre-trained VGGNet (93.78\%). However, with the assistance of unlabeled data, our method can improve the performance by a large margin and achieves accuracy of 91.14\% . Note that the pre-trained ResNet with shortcut connections and bottleneck blocks~\cite{he2016identity} is originally parameter compact, thus when pruning a similar percentage of channels, ResNet is more challenging and usually has larger accuracy drop than that of VGGNet. 

Table \ref{cfvgg} and Table \ref{cfres} also show how the number of labeled images affects the performance of the pruned networks. Fewer data incur larger accuracy drop inevitably, however, the drop of our proposed method is much slower owing to the unlabeled data. For example, with only 100 labeled images, the state-of-the-art Slimming method~\cite{liu2017learning} only achieves 62.49\% accuracy, which is unacceptable for real applications. However, the improvement by unlabeled data are very prominent (\ie~accuracy improved more than 12\% comparing to Slimming~\cite{liu2017learning}). The results show that unlabeled data provide a good platform to transfer the knowledge of the giant network and improve the accuracy accordingly, which is essential when labeled data are extremely limited.

The detailed structure of the pruned VGGNet and ResNet are shown in Table \ref{cifvggstr} and Figure \ref{cifresstr}, respectively.  For VGGNet on CIFAR-10 dataset, more than 90\% channels can be pruned in the later layers, implying  much redundancy. For ResNet with bottleneck structure, a large number of channels in the ``wider'' layers can be pruned. 

\begin{table}[t]
	\caption{Detailed struture of VGGNet on CIFAR-10 dataset by the proposed PUD method. ``\# Channel'' and ``\# Channel*'' denote the number of output channels of convolutional layers in the giant network and the pruned network, respectively.}
	\label{cifvggstr}
	\centering
	\small
	\begin{tabular}{c||ccc} 
		Layer & \# Channel &\# Channel* &Pruning rate (\%)\\ \hline
		conv 1-1&64    & 45    & 29.69 \\
		conv 1-2&64    & 60    & 6.25 \\
		conv 2-1&128   & 120   & 6.25 \\
		conv 2-2&128   & 112   & 12.50 \\
		conv 3-1&256   & 218   & 14.84 \\
		conv 3-2&256   & 211   & 17.58 \\
		conv 3-3&256   & 205   & 19.92 \\
		conv 3-4&256   & 124   & 51.56 \\
		conv 4-1&512   & 64    & 87.50 \\
		conv 4-2&512   & 59    & 88.48 \\
		conv 4-3&512   & 61    & 88.09 \\
		conv 4-4&512   & 37    & 92.77 \\
		conv 4-5&512   & 41    & 92.00 \\
		conv 4-6&512   & 39    & 92.38 \\
		conv 4-7&512   & 44    & 91.41 \\
		conv 4-8&512   & 248   & 51.56 \\ \hline
	\end{tabular}
\end{table}

\begin{table*}[t]	
	\caption{Classification accuracy (top-5) of the pruned VGGNet on ISLVRC2012 dataset with the unlabeled COCO dataset. All the methods achieve approximately 5$\times$ acceleration rate (1.5B FlOPs).}
	\label{resimg}
	\centering
	\small
	\begin{tabular}{c||c|c|c|c|c|c|c}\hline
		\multirow{2}{*}{$N^l$}&\multirow{2}{*}{Scratch}&\multirow{2}{*}{Vanilla Pruning}&SSL&PFEC&Slimming&ISTA-Pruning&\multirow{2}{*}{PUD (Ours)} \\
		
		&&&\cite{wen2016learning}&\cite{li2016pruning}&\cite{liu2017learning} &\cite{ye2018rethinking} &\\ \hline
		50K&65.46&74.76&74.64&74.81&74.96&75.51&\textbf{78.41}\\  \hline
		100K&70.37&76.94&77.12&77.31&77.52&77.84&\textbf{82.21}\\ 
		\hline
	\end{tabular}
\end{table*}

\subsection{Experiments on ImageNet Dataset}
\begin{figure}[t]
	\centering
	\subfigure[ISLVRC2012 dataset.]
	{\includegraphics[width=0.45\columnwidth]{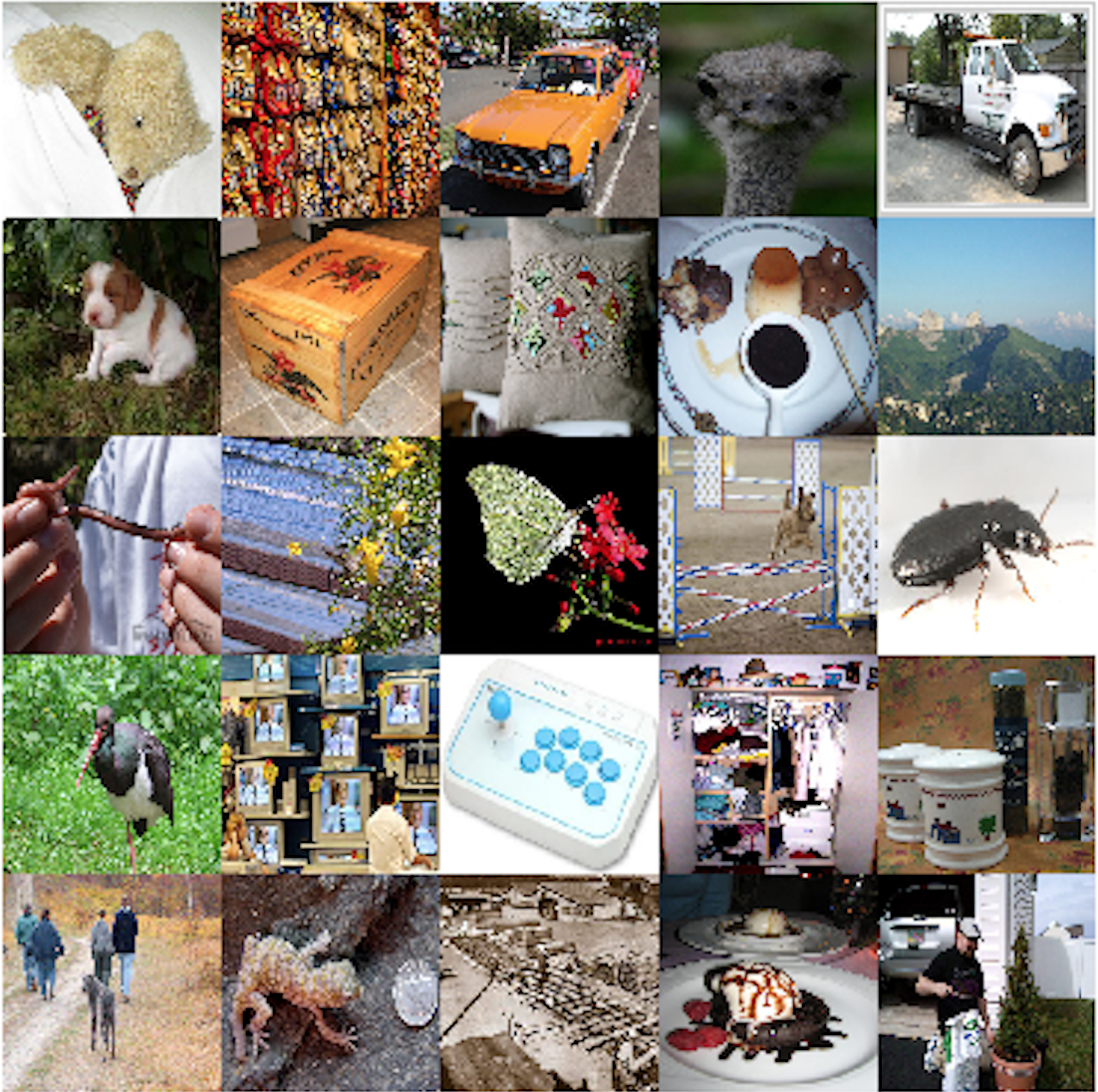}}
	~	\subfigure[COCO dataset.]
	{\includegraphics[width=0.45\columnwidth]{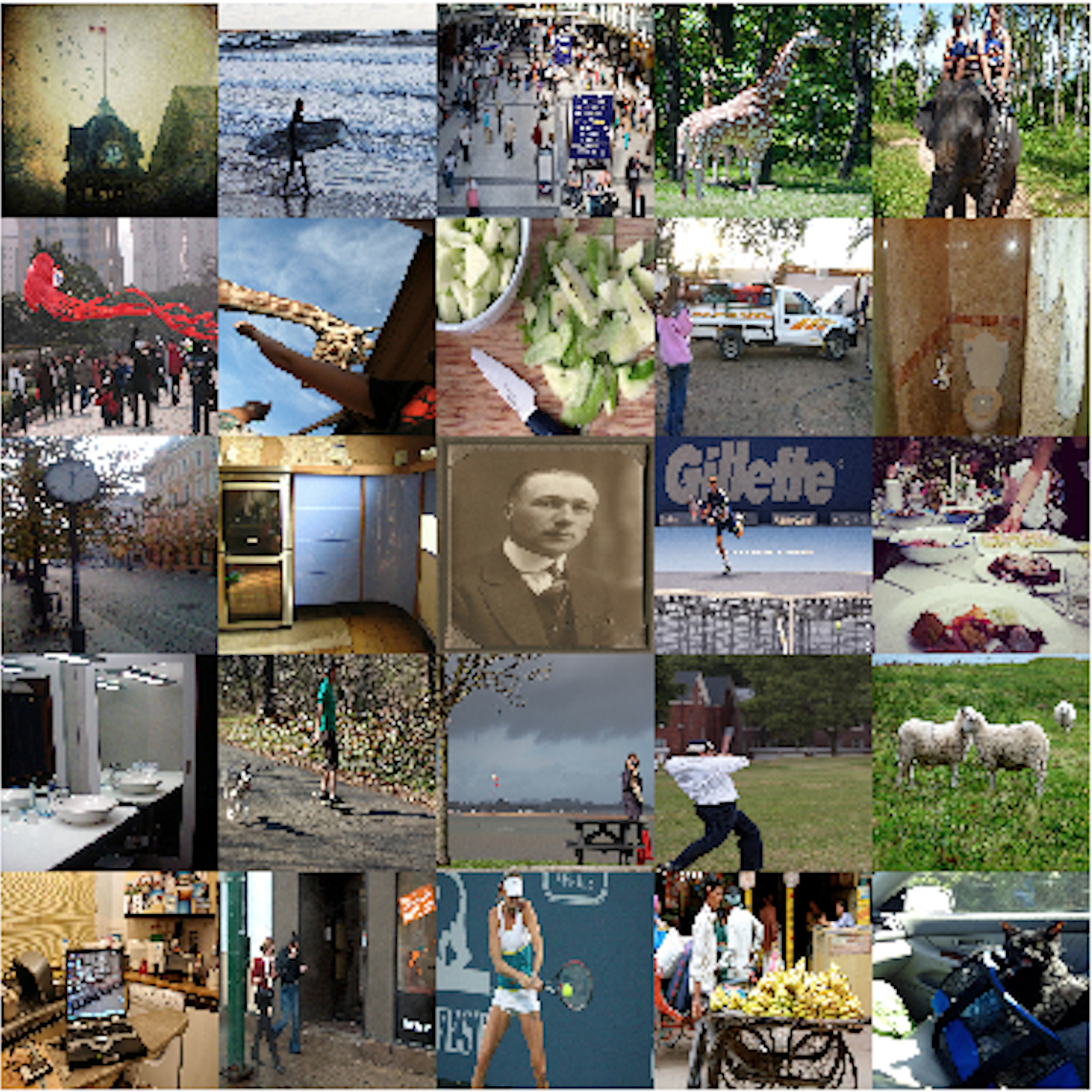}}
	\caption{Sample images in the labeled ISLVRC2012 dataset~\cite{ILSVRC15} and the unlabeled COCO dataset~\cite{LinMBHPRDZ14}.}
	\label{pics:imagenet}
	\vskip -0.1in
\end{figure}

\textbf{Dataset.} The ImageNet (ISLVRC2012) dataset~\cite{ILSVRC15} contains over 1.2M training images and 50k validation images from 1000 categories. For training, all images are cropped with size $224\times 224$ and then randomly horizontally flipped. As for the unlabeled data, COCO dataset~\cite{LinMBHPRDZ14} is adopted since it is also a large-scale benchmark image dataset, which is widely used for object detection, segmentation and captioning. COCO dataset has 80 object categories, much fewer than the ISLVRC2012 dataset. We randomly sample 100k images as the unlabeled data. Using COCO dataset to assist the ISLVRC2012 dataset is a very challenging task because of the large difference of their distributions and categories. Some example images are shown in Figure \ref{pics:imagenet}.

\textbf{Networks.} Following~\cite{liu2017learning}, we use the ``VGG-A'' network model~\cite{simonyan2014very} with batch normalization~\cite{ioffe2015batch} released by PyTorch \footnote{\url{https://pytorch.org/docs/master/torchvision/models.html}} as pre-trained model and evaluate performance with top-5 single-center-crop validation accuracy . The pre-trained model has 89.81\% top-5 accuracy with  7.62B FLOPs. The feature maps after the second pooling layer are sent to a 4-layer convolutional discriminator, which has a similar structure with that for CIFAR-10 dataset, but a convolution layer is added in the beginning. All the convolution layers have $3\times3$ kernels with stride 2.

\textbf{Training.} For all comparison methods, we use 100k iterations for the sparse retraining and 50k iterations for fine-tuning. The initial learning rate is set to 0.01 for the sparse retraining and determined from  $\{0.001,0.003,0.005\}$ for fine-tuning. The learning rate drops by 0.3 at 40\%, 70\% and 90\% of the total iterations. The weights $\alpha$, $\beta$ are respectively set to 0.5, $10^{-6}$ and $\eta$ is selected from $\{0.001,0.01\}$. The sparsity weight $\lambda$ is set to 0.005 for the case $N^l = 50K$ and 0.003 for the case $N^l=100K$. For all methods, we prune 50\% channels of the pre-trained network. 

\textbf{Results.} We randomly sample 50k and 100k labeled images from ISLVRC2012 dataset to implement the compression, assisted with 100K unlabeled samples from the COCO dataset. As Table \ref{resimg} shows, after 50\% channels pruned, all the pruned networks have approximately $5\times$ acceleration rate.  However, the proposed PUD method achieves the best classification accuracies in all cases. It can be safely concluded that the usage of unlabeled data does enable to boost the compression performance on large-scale datasets. Comparing our results with that of ISTA-Pruning, \eg~78.41\% vs 75.51\% and 82.21\% vs 77.84\% for top-5 accuracy, we can infer that the classification ability of the pre-trained model is well preserved by the unlabeled data via mimicking softened output and fixing the dataset bias. Considering the difference between ISLVRC2012 dataset and COCO dataset, the significant improvement on the accuracies shows the effectiveness and superiority of our proposed method. The detailed structure of the pruned VGG-A is shown in Table \ref{imgvggstr}. For the VGG-A on ISLVRC2012 dataset, most of the layers has similar  redundancy.

\begin{table}[t]
	\caption{Detailed structure of the pruned VGG-A model on ISLVRC2012 dataset by the proposed PUD method. ``\# Channel'' and ``\# Channel*'' denote the number of output channels of convolutional layers in the giant network and the pruned network,respectively.}
	\label{imgvggstr}
	\centering
	\small
	\begin{tabular}{c||ccc} 
		Layer & \# Channel &\# Channel* &Pruning rate (\%)\\ \hline
		conv 1-1&64    & 30    & 52.13 \\
		conv 2-1&128   & 57   & 55.47 \\
		conv 3-1&256   & 85   & 66.80 \\
		conv 3-2&256   & 123   & 51.95 \\
		conv 4-1&512   & 172    & 66.41 \\
		conv 4-2&512   & 223    &56.45\\
		conv 4-3&512   & 238    & 53.52 \\
		conv 4-4&512   & 499    & 2.54 \\ \hline
	\end{tabular}
\end{table}

\begin{table*}[t]
	\caption{Effect of each individual component for pruning VGGNet on CIFAR-10 dataset with the unlabeled STL-10 dataset.  $N^u=5K$.}
	\label{abvgg}
	\centering
	\small
	\begin{tabular}{l||cccc|ccc}\hline
		\multirow{7}{*}{VGGNet}&Distillation &Confidence &Adversarial & Rademacher &$N^l=100$&$N^l=500$&$N^l=1K$ \\ \cline{2-8}
		
		&$\times$ &$\times$&$\times$&$\times$  & 62.49  & 84.52 & 87.23\\	
		&$\surd$ &$\times$&$\times$&$\times$ & 70.34&87.62&89.84\\
		&$\surd$ &$\surd$&$\times$&$\times$  & 71.26 &88.18  & 90.21 \\
		&$\surd$ &$\surd$&$\surd$&$\times$ & 72.61  &88.10  &90.72 \\
		&$\surd$ &$\surd$&$\times$&$\surd$ & 73.82 &88.19  &90.50  \\
		
		&$\surd$ &$\surd$&$\surd$&$\surd$ &\textbf{75.04} &\textbf{88.51} &\textbf{91.14}   \\ \hline
	\end{tabular}	
\end{table*}

\begin{table*}[t]
	\caption{Effect of each individual component for pruning ResNet-56 on CIFAR-10 dataset with the unlabeled STL-10 dataset.   $N^u=5K$.}
	\label{abres}
	\centering
	\small
	\begin{tabular}{l||cccc|ccc}\hline
		\multirow{7}{*}{ResNet}&Distillation &Confidence &Adversarial & Rademacher &$N^l=200$&$N^l=500$&$N^l=1K$\\ \cline{2-8}
		
		&$\times$ &$\times$&$\times$&$\times$  & 55.48  & 67.02 & 79.22 \\
		&$\surd$ &$\times$&$\times$&$\times$ &59.31&68.89 &80.97\\
		&$\surd$ &$\surd$&$\times$&$\times$ &59.92&69.37& 81.28\\
		&$\surd$ &$\surd$&$\surd$&$\times$ &60.51 &71.95&82.33 \\
		&$\surd$ &$\surd$&$\times$&$\surd$ &61.18 &72.45& 82.16 \\
		
		&$\surd$ &$\surd$&$\surd$&$\surd$ &\textbf{62.03}&\textbf{73.29}& \textbf{82.42} \\ \hline
	\end{tabular}	
	
\end{table*}
\begin{figure}[t]
	\centering
	\subfigure[]
	{\includegraphics[width=0.46\columnwidth]{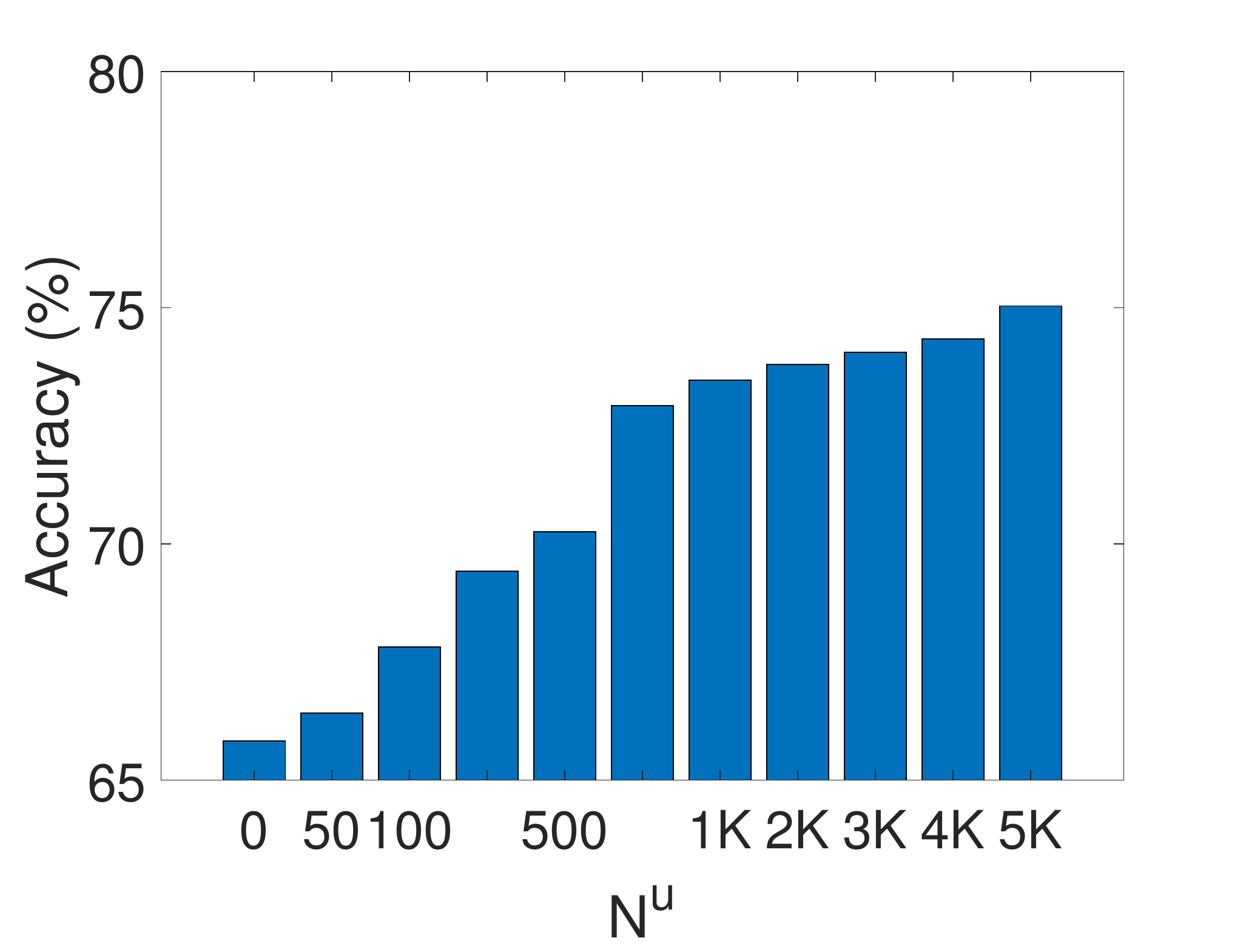}} 
	\subfigure[]
	{\includegraphics[width=0.46\columnwidth]{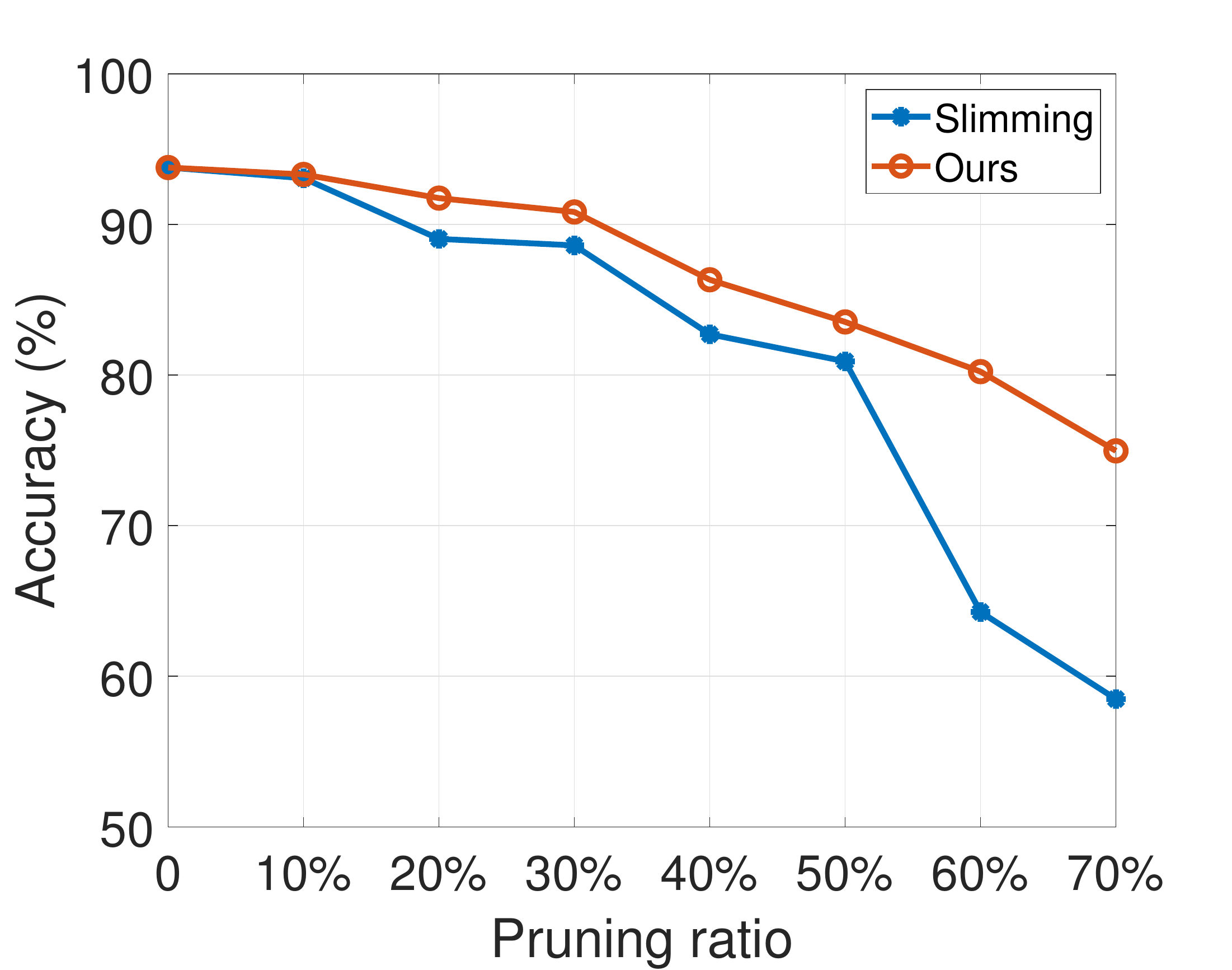}}
	\vskip -0.1in
	\caption{Classification accuracy of the pruned networks on CIFAR-10 dataset \wrt,~(a) different number $N^u$ of unlabeled data with 100 labeled data ($N^l=100$) and 70\% pruning ratio, and (b) various pruning ratio with $N^l=100,~N^u=5K$.}
	\label{analysis_trend}
	\vskip -0.1in
\end{figure}
\begin{figure*}[t]
	\centering
	\subfigure[$\alpha$ for distillation loss $\L_m$.]
	{\includegraphics[width=0.6\columnwidth]{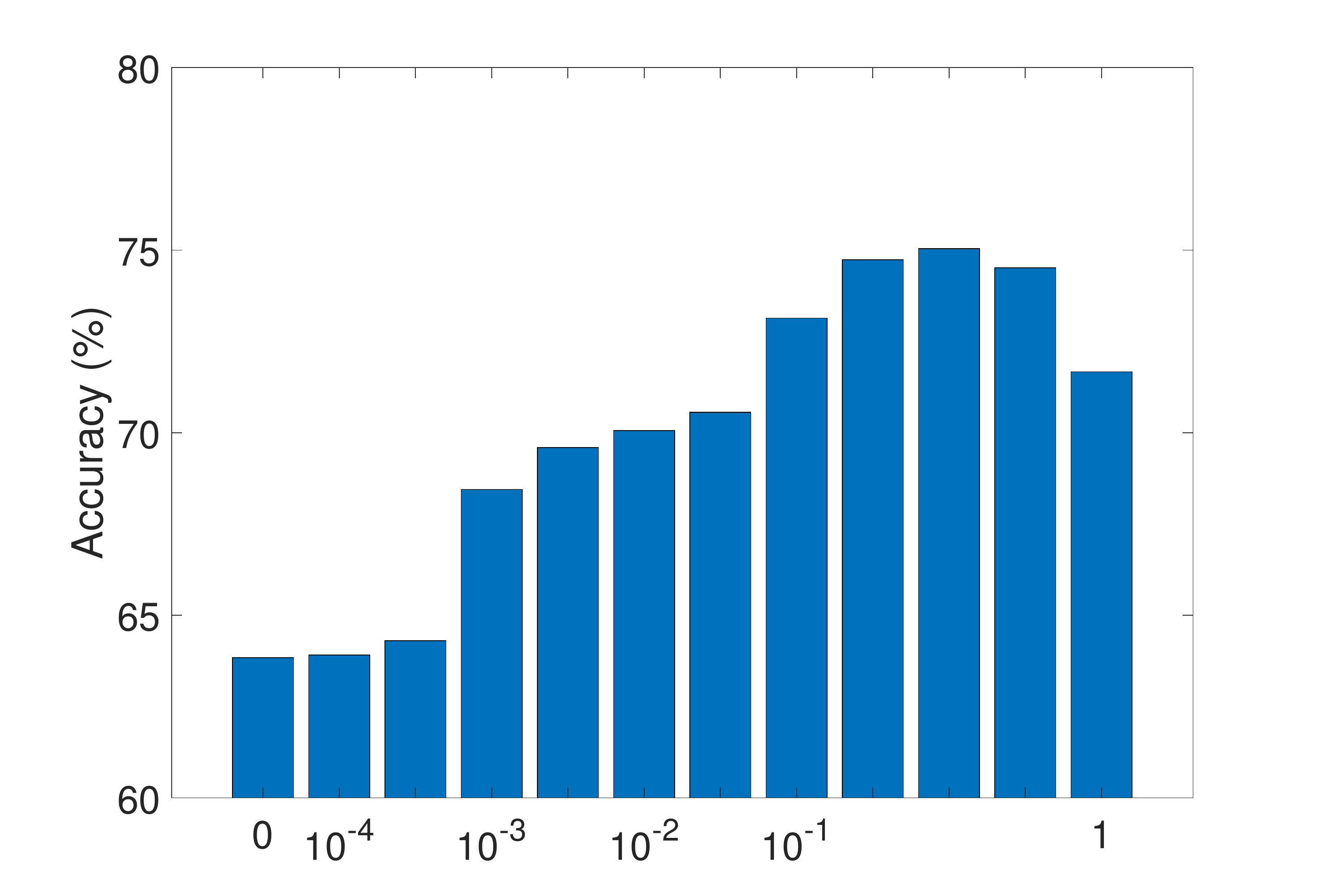}}
	\subfigure[$\beta$ for adversarial loss $\L_{f_1}$.]
	{\includegraphics[width=0.6\columnwidth]{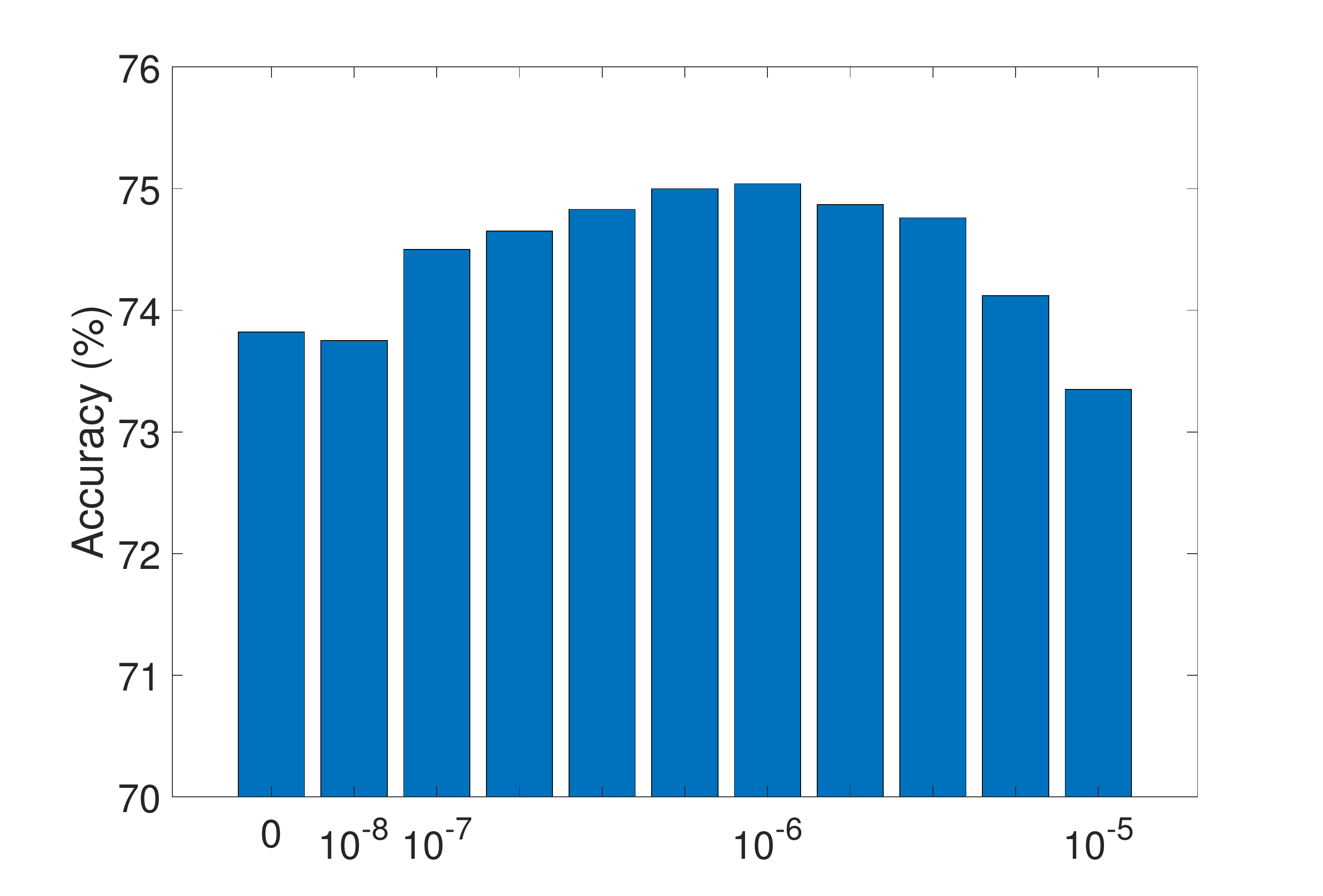}}	
	\subfigure[$\eta$ for Rademacher loss $R_c$.]
	{\includegraphics[width=0.6\columnwidth]{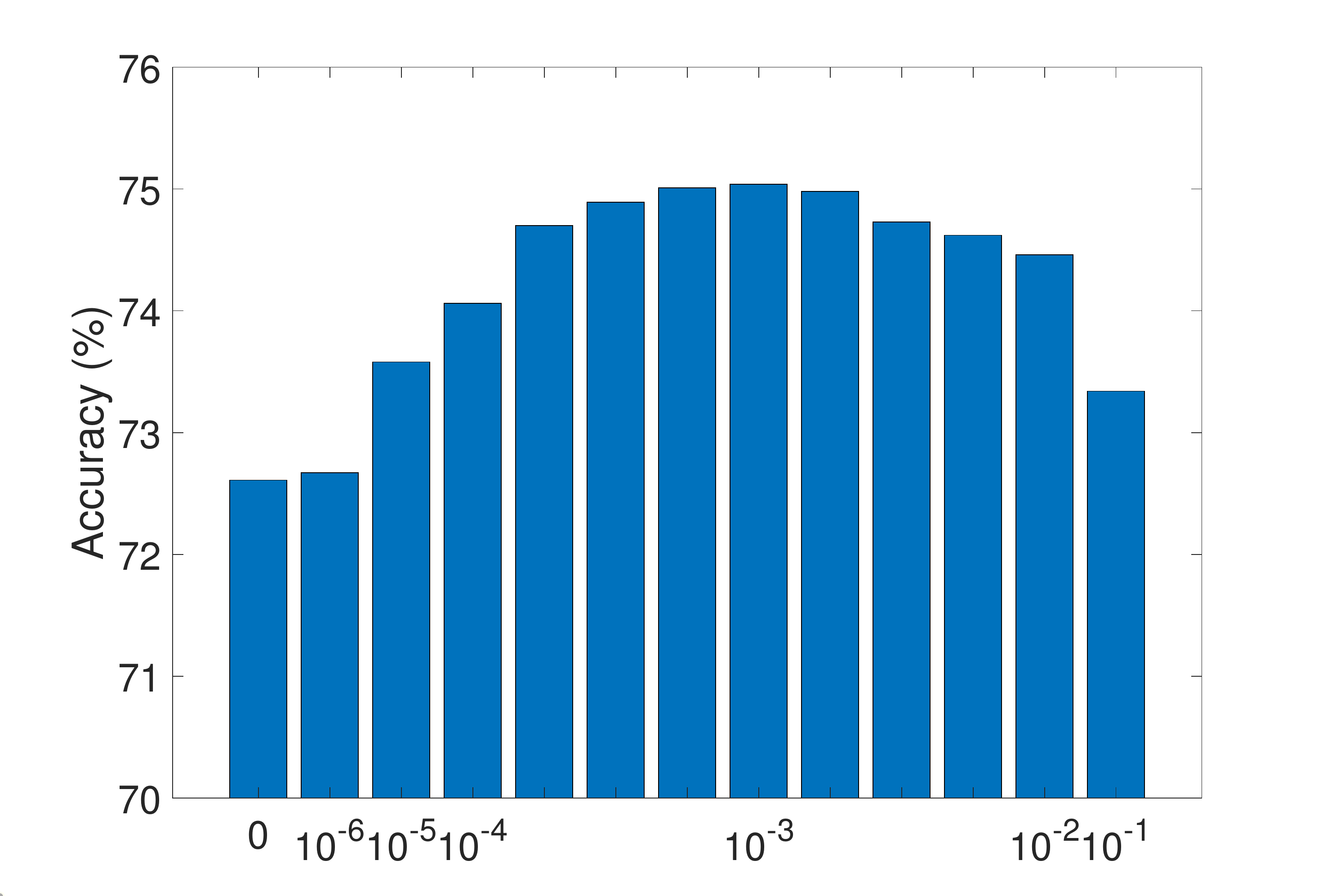}}	
	\caption{Analysis on VGGNet of the three losses in Eq. \eqref{losssum4} by varying their weights. }
	\label{parameters}
\end{figure*}
\subsection{Ablation Studies}
\subsubsection{Effect of the number of unlabeled data}

Furthermore, we investigate how the number of unlabeled data influences the classification accuracy of the pruned networks. In Figure \ref{analysis_trend}(a), we report the corresponding accuracies with 100 labeled examples and various numbers of unlabeled ones. As shown in the results, when the number of unlabeled data are fairly limited, the help is also limited and the accuracy is low accordingly. But with the increase of unlabeled data, the accuracies rise steadily. When the unlabeled data are much more than labeled data, \eg~1K vs 100, the accuracy tends to stabilize. Note that more unlabeled data also bring more training cost, thus in practice for the sake of training efficiency, users do not need to collect too many unlabeled examples. 

\subsubsection{Effect of pruning ratio}
We also investigate how the accuracy of the pruned networks changes when we prune different ratios of their channels. As Figure \ref{analysis_trend}(b) shows, drop of accuracy occurs as more channels are pruned since more information stored in the giant network loses and cannot be recovered totally due to limited labeled data. The accuracy of the proposed PUD method is always higher than that of pruning without unlabeled data, especially for a high pruning ratio (\eg~60\%). This might be because our method can leverage the unlabeled data to decrease the loss of information in the sparse retraining and restore information in fine-tuning as well. 

\subsubsection{Effect of each individual component}
We now investigate the effect of each individual component in the proposed PUD method, \ie~the distillation loss  $\mathcal{L}_m$, the adversarial loss $\L_{f_1}$, Rademacher loss $R_c$ and confidence on unlabeled data $C_i^u$. The accuracies with (``$\surd$'') or without  (``$\times$'') each component for both VGGNet and ResNet on CIFAR-10 are shown in  Table \ref{abvgg} and Table \ref{abres}. Distillation loss directly involves unlabeled data in the retraining process, and improves the performance by a large margin (\ie~from 87.23\% to 89.84\%), which verifies the prominent effect of unlabeled data as a good platform to distill knowledge from the pretrained network,  as well as further alleviate over-fitting. Weighting unlabeled data with confidence $C_i^u$ further improves performance (\ie~from 89.94\% to 90.21\%). However, due to the bias between labeled and unlabeled data, there is still a large room to boost performance. The adversarial loss $L_{f_1}$ alleviates the bias in low-level feature space, making unlabeled data have more positive effect and resulting in the improvement from 90.21\% to 90.72\%. Loss $R_c$ derived from the theoretical generalization error bound strengthens the robustness of the proposed method as well as  enhances performance slightly (\ie~from 90.21\% to 90.50\%). With all the components and their mutual effect, the proposed method achieves the best performance (\ie~91.14\%). 


To further study how each individual loss and its corresponding weight coefficient affect the final performance, we vary weights $\alpha$, $\beta$ and $\eta$ by fixing the others at the optimal parameter configuration  with 100 labeled images and 5K unlabeled data, as shown in Figure \ref{parameters}.

\textbf{Distillation loss $\mathcal{L}_m$ and weight $\alpha$.} The main function of loss $\L_m$ is to encourage the spare network to mimic the classification characteristics of the pre-trained model on unlabeled data. When varying $\alpha$ from 0 to 1, the degree of distillation increases accordingly. From Figure \ref{parameters}(a), the accuracy achieves a high level when $\alpha$ exceeds 0.001 then increases steadily with $\alpha$. We also observe that an overlarge $\alpha$ (\eg~1) would induce the accuracy to drop a bit. This might result from that in practice, the distribution of unlabeled data is different from that of labeled data, and an overemphasis on the unlabeled data would disturb the network's fitting ability on the labeled data. 

\textbf{Adversarial loss $\L_{f_1}$ and weight $\beta$.} The low-level features on the pre-trained model are usually fairly different between unlabeled and labeled data, thus the adversarial loss is very large at the beginning of the retraining. We empirically find a small $\beta$ can still have a more significant impact on the update of low-level features than that on the output layer since the adversarial loss is directly imposed on the low-level layers.  The adversarial loss aligns the distributions of the unlabeled and labeled low-level features; however, this might cause that the feature distributions of labeled data on the pruned network drift away slightly from that on the giant network. In Figure \ref{parameters}(b), when the weight $\beta$ is too large, much information of the original giant network will lose and the accuracy drops slightly.
\begin{table}[t]
	\caption{Effect of labeled data and unlabeled data for the distillation loss $L_m$.}
	\label{mimlu}
	\centering
	\small
	\begin{tabular}{cc|ccc}\hline
		L&U &$N^l=100$&$N^l=500$&$N^l=1K$ \\ \hline
		$\times$ & $\times$& 63.57  &85.26 & 87.86 \\
		$\surd$ &$\times$&65.83 & 85.89&88.14 \\  
		$\times$ & $\surd$&74.71 &88.44&91.05\\
		$\surd$ & $\surd$&75.04&88.51& 91.14\\\hline
	\end{tabular}	
\end{table}

\begin{table}[t]
	\caption{Effect of labeled data and unlabeled data for the Rademacher loss $R_c$.}
	\label{radlu}
	\centering
	\small
	\begin{tabular}{cc|ccc}\hline
		L&U &$N^l=100$&$N^l=500$&$N^l=1K$ \\ \hline
		$\times$ & $\times$& 72.61  & 88.10 &90.72 \\ 
		$\surd$ &$\times$& 72.66&88.11 &90.95 \\
		$\times$ & $\surd$& 74.53&88.42&91.10 \\
		$\surd$ & $\surd$&75.04 &88.51&91.14\\\hline
	\end{tabular}	
\end{table}


\textbf{Rademacher loss $R_c$ and weight $\eta$.} Rademacher loss acts as a regularization term to boost the generalization ability of the pruned networks. The Rademacher loss complements with the two losses $\L_u$ and $\L_{f_1}$, and work best with $\eta=0.001$.  Nevertheless, stronger regularization (\eg~0.1 in Figure \ref{parameters}(c)) may also hamper the classification accuracy. 

\textbf{Verification of limitation of few labeled data.} Note that the distillation loss $\L_m$ and Rademacher loss $R_c$ can be imposed both on labeled data and unlabeled data. By fixing other components, we additionally conduct experiments whether the implementation of $\L_m$ and $R_c$ cover the labeled data or unlabeled data. When implementing $\L_m$ (or $R_c$) on both labeled and unlabeled data we try different weights on them, which has subtle affect due to the limitation of labeled data, and thus equal weights are used for simplicity.   Accuracies of the pruned VGGNet on CIFAR-10 are presented in Table \ref{mimlu} and Table \ref{radlu}, and "L" represents labeled data while ``U'' is for unlabeled data. We can see that for both distillation loss and Rademacher loss, implementing them only with labeled data has a small effect. For example, with 100 labeled data for distillation loss (Rademacher loss), the improvement of performance is only 2.26\% (0.05\%). However, when introducing unlabeled data, the performance can be improved for a large margin. Even implementing distillation loss (Rademacher loss) only with unlabeled data, the accuracy is improved by 10.84\% (1.92\%) accordingly. We can safely conclude that the unlabeled data do play a vital part in helping the performance improvement of pruned networks.

\begin{table}[t]
	\caption{Comparison with state-of-the-art knowledge distillation methods.}
	\label{otkd}
	\centering
	\small
	\begin{tabular}{c|ccc}\hline
		Method &$N^l=100$&$N^l=500$&$N^l=1K$ \\ \hline
		W/o unlabeled data&62.49&84.52&87.23\\
		KD~\cite{hinton2015distilling} &70.34&87.62&89.84\\
		AT~\cite{zagoruyko2016paying} & 70.78 & 87.73 & 90.06\\ 
		NST~\cite{huang2017like} & 71.23&87.76 &90.25 \\
		Ours &75.04 &88.51&91.14\\\hline
	\end{tabular}	
\end{table}

\textbf{Comparison with state-of-the-art knowledge distillation methods.} For utilizing unlabeled data to assist compression, several existing knowledge distillation methods can be also directly applied, which pushes the spare network to behave similar as the giant network on those unlabeled data. We compare the proposed method with several state-of-the-art knowledge distillation methods and the accuracies of the pruned VGGNet on CIFAR-10 are shown in Table~\ref{otkd}. Unlabeled data can improve the performance of the pruned network with different distillation manners, implying the effectiveness of unlabeled data. However, the unlabeled data may be different from  labeled data in many aspects such as distribution and categories, which limits the potentiality of exploring unlabeled data and the inappropriate data may disturb the training procedure. The proposed method alleviates dataset bias by accommodating the unlabeled data to the pre-trained network  and then uses the instance-wise confidences to weight different unlabeled data, which further improves the performance dramatically.      

\section{Conclusion}
\label{sec:con}

We solved a practical problem of compressing giant neural networks given only a few labeled examples instead of the original and complete training dataset. We exploited the unlabeled data to distill the knowledge from the giant network into the pruned network and boosted the compression performance. To alleviate the dataset bias between labeled and unlabeled data, we utilized the low-level layers of the network as an aligner to make an alignment on their representations. Experimental results validated the effectiveness of the proposed PUD method. For the future work, we plan to investigate an extreme situation even if no single example is released with the giant networks, which might demand higher generalization ability of the compressed networks.


%

\ifCLASSOPTIONcaptionsoff
\newpage
\fi



%
%
%
\bibliographystyle{IEEEtran}
\bibliography{reference}

%

\begin{IEEEbiography}[{\includegraphics[width=1in,height=1.25in,clip,keepaspectratio]{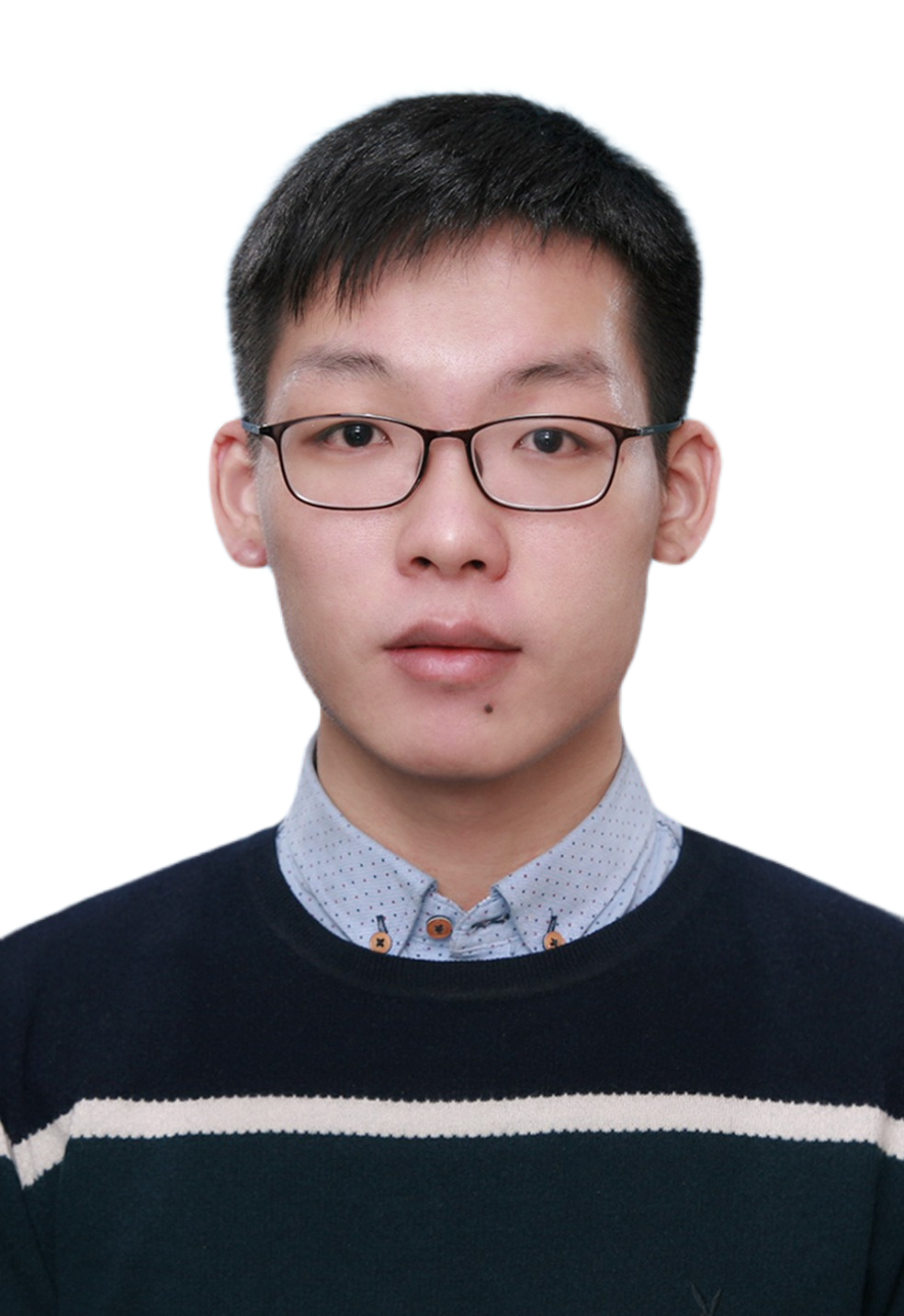}}]{Yehui Tang}
	received the B.E. degree from Xidian University in 2018. Currently, he is a Ph.D candidate with the Key Laboratory of Machine Perception (Ministry of Education) at Peking University. His research interests lie primarily in machine learning and computer vision.
\end{IEEEbiography}

\begin{IEEEbiography}[{\includegraphics[width=1in,height=1.25in,clip,keepaspectratio]{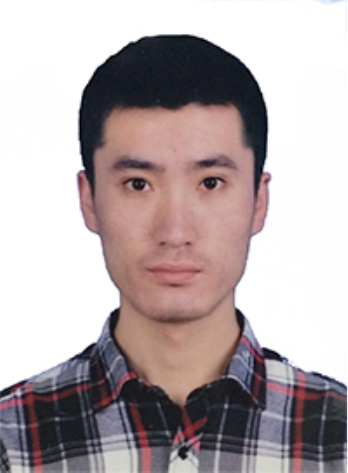}}]{Shan You} is now a researcher at SenseTime Research. He obtained the B.E degree from Xi'an Jiaotong University in 2014, and a Ph.D degree with the Key Laboratory of Machine Perception (Ministry of Education) at Peking University. His research interests lie primarily in machine learning and computer vision.
\end{IEEEbiography}

\begin{IEEEbiography}[{\includegraphics[width=1in,height=1.25in,clip,keepaspectratio]{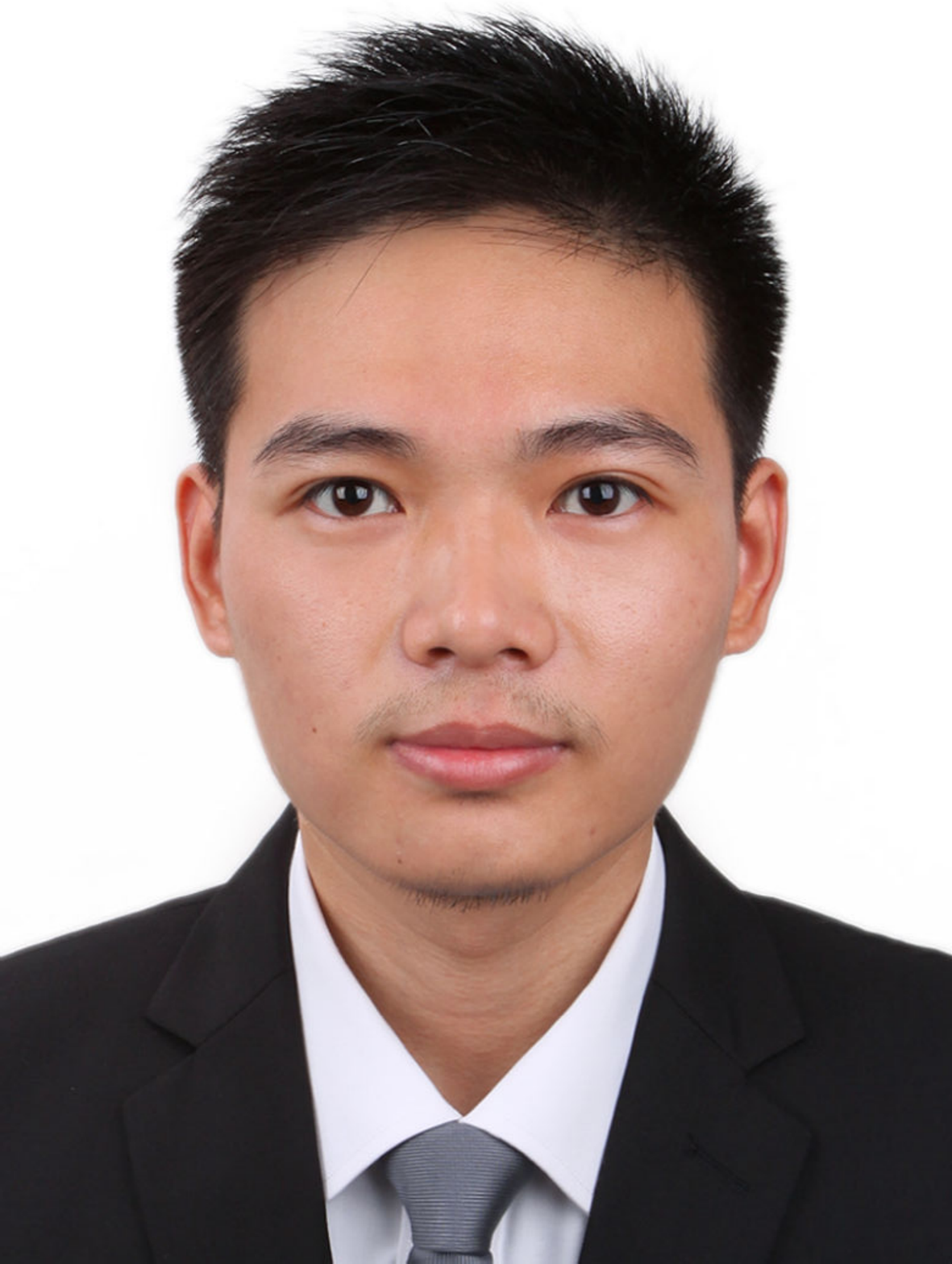}}]{Chang Xu}
	is a Lecturer in Machine Learning and Computer Vision at the School of Information Technologies, The University of Sydney. He obtained a Bachelor of Engineering from Tianjin University, China, and a Ph.D. degree from Peking University, China. While pursing his PhD degree, Chang received fellowships from IBM and Baidu. His research interests lie in machine learning, data mining algorithms and related applications in artificial intelligence and computer vision, including multi-view learning, multi-label learning, visual search and face recognition. His research outcomes have been widely published in prestigious journals and top tier conferences.
\end{IEEEbiography}
\begin{IEEEbiography}
	[{\includegraphics[width=1in,height=1.25in,clip,keepaspectratio]{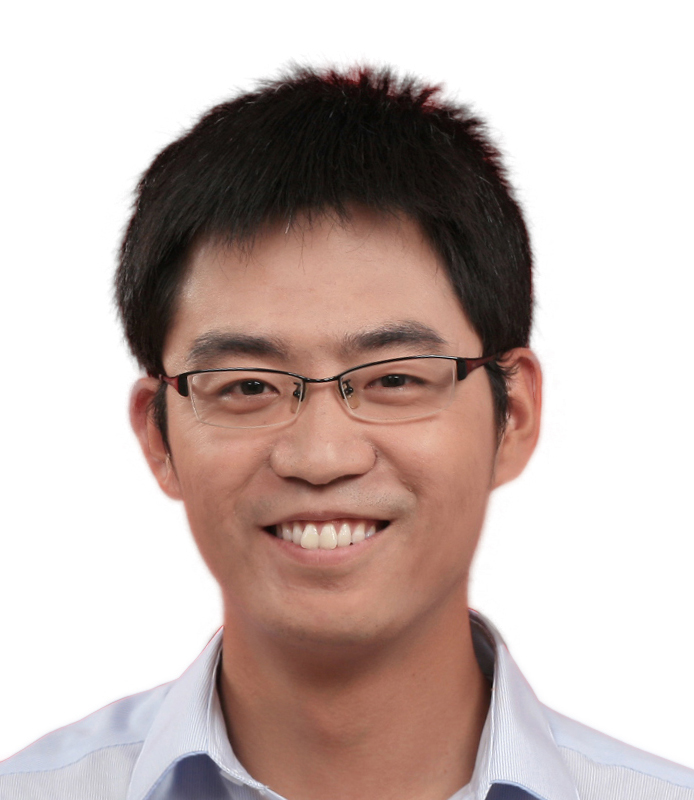}}] {Boxin Shi} 
	is currently a Boya Young Fellow Assistant Professor at Peking University, where he leads the Camera Intelligence Group. Before joining PKU, he did postdoctoral research at MIT Media Lab, Singapore University of Technology and Design, Nanyang Technological University from 2013 to 2016, and worked as a Researcher at the National Institute of Advanced Industrial Science and Technology from 2016 to 2017. He received the B.E. degree from Beijing University of Posts and Telecommunications in 2007, M.E. degree from Peking University in 2010, and Ph.D. degree from the University of Tokyo in 2013. He won the Best Paper Runner-up award at International Conference on Computational Photography 2015. He has served as Area Chairs for ACCV 2018, BMVC 2019, and 3DV 2019.
\end{IEEEbiography}

\begin{IEEEbiography}[{\includegraphics[width=1in,height=1.25in,clip,keepaspectratio]{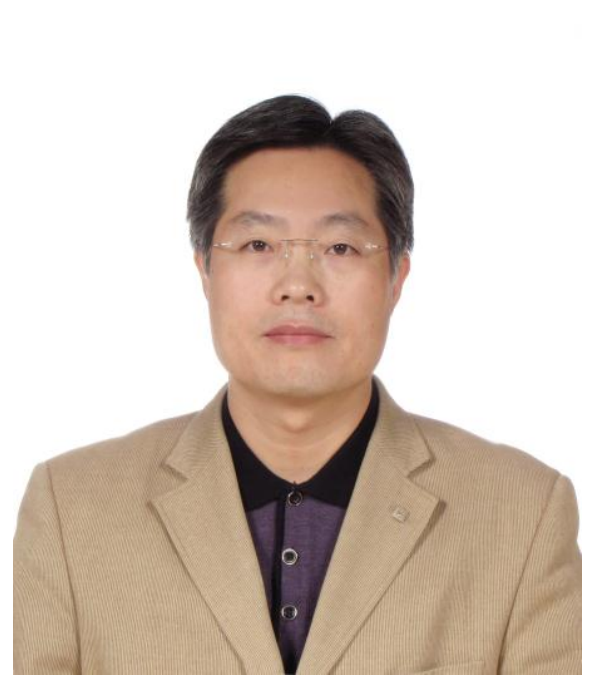}}]{Chao Xu}
	received the B.E. degree from Tsinghua University in 1988, the M.S. degree from University of Science and Technology of China in 1991 and the Ph.D degree from Institute of Electronics, Chinese Academy of Sciences in 1997. Between 1991 and 1994 he was employed as an assistant professor by University of Science and Technology of China. Since 1997 Dr. Xu has been with School of EECS at Peking University where he is currently a Professor. His research interests are in image and video coding, processing and understanding. He has authored or co-authored more than 150 publications and 5 patents in these fields.	
\end{IEEEbiography}

\end{document}